\pdfoutput=1

\documentclass[11pt]{article}

\usepackage{acl}

\usepackage{times}
\usepackage{latexsym}
\usepackage{caption}
\usepackage{subcaption}
\usepackage{newfloat}
\usepackage{listings}
\usepackage{graphicx}
\usepackage{enumitem}

\usepackage[T1]{fontenc}

\usepackage[utf8]{inputenc}

\usepackage{microtype}

\title{Semantic Segmentation of Legal Documents via Rhetorical Roles}

\author{Vijit Malik$^{1}$\thanks{\ \ Equal Contributions} \qquad Rishabh Sanjay$^{1}$\footnotemark[1] \qquad \textbf{Shouvik Kumar Guha}$^{2}$ \\
\textbf{Angshuman Hazarika}$^{3}$ \qquad \textbf{Shubham Nigam}$^{1}$   \qquad \textbf{Arnab Bhattacharya}$^{1}$\\
\textbf{Ashutosh Modi}$^{1}$\thanks{\ \ Corresponding Author} \\
 $^{1}$Indian Institute of Technology Kanpur (IIT-K) \\\
 $^{2}$West Bengal National University of Juridical Sciences (WBNUJS)\\
 $^{3}$Indian Institute of Management Ranchi (IIM-R)\\
 \texttt{\{vijitvm21,rishabh.lfs\}@gamil.com} \qquad \texttt{snigam@cse.iitk.ac.in} \\ 
 \texttt{shouvikkumarguha@nujs.edu}  \qquad
\texttt{angshuman.hazarika@iimranchi.ac.in} \\
 \texttt{\{arnabb,ashutoshm\}@cse.iitk.ac.in}\\ 
}

\begin{document}
\maketitle
\begin{abstract}
Legal documents are unstructured, use legal jargon, and have considerable length, making them difficult to process automatically via conventional text processing techniques. A legal document processing system would benefit substantially if the documents could be segmented into coherent information units. This paper proposes a new corpus of legal documents annotated (with the help of legal experts) with a set of 13 semantically coherent units labels (referred to as Rhetorical Roles), e.g., facts, arguments, statute, issue, precedent, ruling, and ratio. We perform a thorough analysis of the corpus and the annotations. For automatically segmenting the legal documents, we experiment with the task of rhetorical role prediction: given a document, predict the text segments corresponding to various roles. Using the created corpus, we experiment  extensively with various deep learning-based baseline models for the task. Further, we develop a multitask learning (MTL) based deep model with document rhetorical role label shift as an auxiliary task for segmenting a legal document. The proposed model shows superior performance over the existing models. We also experiment with model performance in the case of domain transfer and model distillation techniques to see the model performance in limited data conditions. 
\end{abstract}

\section{Introduction} \label{sec:intro}


The number of legal cases has been growing almost exponentially in populous countries like India. For example, as per the India's National Judicial Data Grid, there are about 41 million cases pending in India \cite{njdc-district}. As per some of recent estimates by a retired Supreme Court of India Judge, it will take about 450 years to clear the backlog of cases \cite{backlogcases2019}. Technology could come to the rescue in dealing with the backlog, for example, if there were a technology (based on NLP techniques) that could help a legal practitioner to extract relevant information from legal documents then it could make the legal process more streamlined and efficient. However, legal documents are quite different from conventional documents used to train NLP systems (e.g., newspaper texts). Legal documents are typically long (tens of pages) \cite{malik-etal-2021-ildc}, unstructured \cite{9679940,10.1007/978-3-030-33220-4_20}, noisy (e.g., grammatical and spelling mistakes due to manual typing in courts) \cite{malik-etal-2021-ildc,kapoor-etal-2022-hldc}, and use different lexicon (legal jargon). The use of a specialized lexicon and different semantics of words makes pre-trained neural models (e.g., transformer-based models) ineffective \cite{chalkidis-etal-2020-legal}. The legal domain has several sub-domains (corresponding to different laws, e.g., criminal law, income tax law) within it. Although some of the fundamental legal principles are common, the overlap between different sub-domains is low; hence systems developed on one law (e.g., income tax law) may not directly work for another law (e.g., criminal law), so there is the problem of a domain shift \cite{bhattacharya2019identification,malik-etal-2021-ildc,kalamkar-EtAl:2022:LREC,kapoor-etal-2022-hldc}.

In this paper, we target legal case proceedings in the form of judgment documents. To aid the processing of long legal documents, we propose a method of segmenting a legal document into coherent information units referred to as \textit{Rhetorical Roles} \cite{saravanan2008automatic,bhattacharya2019identification}. We propose a corpus of legal documents annotated with Rhetorical Roles (RRs). RRs could be useful for various legal applications. Legal documents are fairly long, and dividing these into rhetorical role units can help summarize documents effectively. In the task of legal judgment prediction, for example, using RRs, one could extract the relevant portions of the case that contributes towards the final decision. 
RRs could be useful for legal information extraction, e.g., it can help extract cases with similar facts. Similarly, prior cases similar to a given case could be retrieved by comparing different rhetorical role units. In this work, we make the following contributions: 

\noindent 1. We create a new corpus of legal documents annotated with rhetorical role labels. 
In contrast to previous work (8 RRs) \cite{bhattacharya2019identification}, we create a more fine-grained set of 13 RRs. Further, we create the corpus on different legal domains (\S \ref{sec:corpus}).

\noindent 2. For automatically segmenting the legal documents, we experiment with the task of rhetorical role prediction: given a document, predict the text segments corresponding to various roles. Using the created corpus, we experiment with various deep text classification and baseline models for the task. We propose new multi-task learning (MTL) based deep model with document level rhetorical role shift as an auxiliary task for segmenting the document into rhetorical role units (\S \ref{sec:models}). The proposed model performs better than the existing models for RR prediction. We further show that our method is robust against domain transfer to other legal sub-domains (\S \ref{sec:experiments}). We release the corpus, model implementations and experiments code: \url{https://github.com/Exploration-Lab/Rhetorical-Roles}

\noindent 3. Given that annotating legal documents with RR is a tedious process, we perform model distillation experiments with the proposed MTL model and attempt to leverage unlabeled data to enhance the performance (\S \ref{sec:experiments}). We also show the use-case for RR prediction model. 

\section{Related Work} \label{sec:related} 

Legal text processing has been an active area of research in recent times. A number of datasets, applications, and tasks have been proposed. For example, Argument Mining \cite{wyner2010approaches},  
Information Extraction and Retrieval \cite{tran2019building}, Event Extraction \cite{lagos2010event}, Prior Case Retrieval \cite{jackson2003information}, Summarization \cite{moens1999abstracting}, and Case Prediction \cite{malik-etal-2021-ildc, chalkidis-etal-2019-neural, strickson2020legal, kapoor-etal-2022-hldc}. Recently, there has been a rapid growth in the development of NLP and ML technologies for the Chinese legal system, inter alia, \citet{chen-etal-2019-charge, hu-etal-2018-shot, jiang-etal-2018-interpretable, ijcai2019-567, ye-etal-2018-interpretable}. Few works have focused on the creation of annotated corpora and the task of automatic rhetorical role labeling. \citet{venturi2012design} developed a corpus, TEMIS of 504 sentences annotated both syntactically and semantically. The work of \citet{wyner2013case} focuses on the process of annotation and conducting inter-annotator studies. \citet{savelka2018segmenting} conducted document segmentation of U.S. court documents using Conditional Random Fields (CRF) with handcrafted features to segment the documents into functional and issue-specific parts. Automatic labeling of rhetorical roles was first conducted in \citet{saravanan2008automatic}, where CRFs were used to label seven rhetorical roles. \citet{nejadgholi2017semi} developed a method for identification of factual and non-factual sentences using fastText. The automatic ML approaches and rule-based scripts for rhetorical role identification were compared in \citet{walker2019automatic}. \citet{kalamkar-etal-2022-corpus} create a large corpus of RRs and propose transformer based baseline models for RR prediction. Our work comes close to work by \citet{bhattacharya2019identification}, where they use the BiLSTM-CRF model with sent2vec features to label rhetorical roles in Indian Supreme Court documents. In contrast, we develop a multi-task learning (MTL) based model for RR prediction that outperforms the system of \citet{bhattacharya2019identification}.

\section{Rhetorical Roles Corpus} \label{sec:corpus}

\noindent \textbf{Corpus Acquisition:} We focus on Indian legal documents in English; however, techniques we develop can be generalized to other legal systems.  We consider legal judgments from the Supreme Court of India, High Courts, and Tribunal courts crawled from the website of IndianKanoon ({\url{https://indiankanoon.org/}}). We also scrape Competition Law documents from Indian Tribunal court cases (National Company Law Appellate Tribunal (NCLAT), COMPetition Appellate Tribunal (COMPAT), Competition Commission of India (CCI)). We focus on two domains of the Indian legal system: Competition Law (CL) (also called as Anti-Trust Law in the US and Anti-Monopoly law in China) and Income Tax (IT). CL deals with regulating the conduct of companies, particularly concerning competition. With the help of legal experts, we narrowed down the cases pertinent to CL and IT from the crawled  corpus (also see Ethical Considerations in App. \ref{app:ethics}). 




\noindent \textbf{Choice of CL and IT domains}: India has a common law system where a decision may not be exactly as per the statutes, but the judiciary may come up with its interpretation and overrule existing precedents. This introduces a bit of subjectivity. One of the biggest problems faced during the task of identifying the rhetorical roles in a judgment is that the element of subjectivity involved in the judicial perception and interpretation of different rhetorical roles, ranging from the factual matrix (i.e., perception about facts, relevant facts and facts in an issue may vary) to the statutory applicability and interpretation to determine the fitness of a particular judicial precedent to the case at hand. In order to overcome this particular obstacle, we focus on specific legal domains (CL and IT) that display a relatively greater degree of consistency and objectivity in terms of judicial reliance on statutory provisions to reach decisions \cite{taxmann2021}. 


\noindent \textbf{Corpus Statistics:} We randomly selected a set of 50 documents each for CL and IT from the set of acquired documents ($\approx$ 1.6k for IT and $\approx$ 0.8k for CL). These 100 documents were annotated with 13 fine-grained RR labels (vs. 8 by \citet{bhattacharya2019identification}) by a team of legal experts. Our corpus is double the size of the RR corpus of  \citet{bhattacharya2019identification}. The CL documents have 13,328 sentences (avg. of 266 per document), and IT has a total of 7856 sentences (avg. of 157 per document). Label-wise distribution for IT and CL documents are  provided in Appendix \ref{app:roles}.  
Annotating legal documents with RRs is a tedious as well as challenging task. Nevertheless, this is a growing corpus, and we plan to add more annotated documents. However, given the complexity of annotations, the RR labeling task also points towards looking for model distillation (\S \ref{sec:experiments}) and zero-shot learning-based methods. 


\noindent \textbf{Annotation Setup:} The annotation team (legal team) consisted of two law professors from prestigious law schools and six graduate-level law student researchers. Annotating just 100 documents took almost three months. Based on detailed discussions with the legal team, we initially arrived at the eight main rhetorical roles (facts, arguments, statues, dissent, precedent, ruling by lower court, ratio and ruling by present court) plus one `none' label. During the annotation, roles were further refined, and the documents were finally annotated with 13 fine-grained labels since some of the main roles could be sub-divided into more fine-grained classes. The list of RRs is as follows (example sentences for each role is in Table \ref{app:tab:rr-examples} in the Appendix \ref{app:roles}): 

\begin{itemize}[noitemsep,nolistsep]
    \item \textbf{Fact (FAC):} These are the facts specific to the case based on which the arguments have been made and judgment has been issued. In addition to Fact, we also have the fine-grained label \textbf{Issues (ISS)}. The issues which have been framed/accepted by the present court for adjudication.
    \item \textbf{Argument (ARG)}: The arguments in the case were divided in two more fine-grained sub-labels: \textbf{Argument Petitioner (ARG-P):} Arguments which have been put forward by the petitioner/appellant in the case before the present court and by the same party in lower courts (where it may have been petitioner/respondent). Also, \textbf{Argument Respondent (ARG-R):} Arguments which have been put forward by the respondent in the case before the present court and by the same party in lower courts (where it may have been petitioner/respondent)

    \item \textbf{Statute (STA):} The laws referred in the case.
    \item \textbf{Dissent (DIS):} Any dissenting opinion expressed by a judge in the present judgment/decision.
    
    \item \textbf{Precedent (PRE):} The precedents in the documents were divided into 3 finer labels, \textbf{Precedent Relied Upon (PRE-R):} The precedents which have been relied upon by the present court for adjudication. These may or may not have been raised by the advocates of the parties and amicus curiae. \textbf{Precedent Not Relied Upon (PRE-NR):} The precedents which have not been relied upon by the present court for adjudication. These may have been raised by the advocates of the parties and amicus curiae. \textbf{Precedent Overruled (PRE-O):} Any precedents (past cases) on the same issue which have been overruled through the current judgment.

    \item \textbf{Ruling By Lower Court (RLC):} Decisions of the lower courts which dealt with the same case.
    \item \textbf{Ratio Of The Decision (ROD):} The principle which has been established by the current judgment/decision which can be used in future cases. Does not include the obiter dicta which is based on observations applicable to the specific case only.
    \item \textbf{Ruling By Present Court (RPC):} The decision of the court on the issues  which have been framed/accepted by the present court for adjudication.
    \item \textbf{None (NON):} any other matter in the judgment which does not fall in any of the above-mentioned categories.
\end{itemize}

The dataset was annotated by six legal experts (graduate law student researchers), 3 annotated 50 CL documents, and the remaining 3 annotated 50 IT documents. We used Webanno \cite{de2016web} as the annotation framework. Each legal expert assigned one of the 13 Rhetorical roles to each document sentence. Note that we initially experimented with different levels of granularity (e.g., phrase level, paragraph level), and based on the pilot study, we decided to go for sentence-level annotations as it maintains the balance (from the perspective of topical coherence) between too short (having no labels) and too long (having too many labels) texts. Legal experts pointed out that a single sentence can sometimes represent multiple rhetorical roles (although this is not common). Each expert could also assign secondary and tertiary rhetorical roles to a single sentence to handle such scenarios (also App. \ref{app:secondary-tertiary}). As an example, suppose a sentence is a `Fact' but could also be an `Argument' according to the legal expert. In that case, the expert could assign the rhetorical roles `Primary Fact' and `Secondary Argument' to that sentence. We extended it to the tertiary level as well to handle rare cases.   

Our corpus is different from the existing corpus \cite{bhattacharya2019identification}. Firstly, we use 13 fine-grained RR labels and the size of the corpus is almost twice. Secondly, we focus on different legal sub-domains (IT and CL vs. Supreme Court Judgments). Lastly, we perform the primary, secondary, and tertiary levels of annotations since, according to legal experts, it is sometimes possible that a sentence might have multiple RR labels.

\begin{table}[t]
\small
\centering
\begin{tabular}{|l|l|l|}
\hline
\textbf{Label}       & \textbf{IT}   & \textbf{CL}    \\ \hline
\textbf{AR}          & 0.80         & 0.93          \\ \hline
\textbf{FAC}         & 0.80         & 0.89          \\ \hline
\textbf{PR}          & 0.70           & 0.86           \\ \hline
\textbf{STA}         & 0.78         & 0.89          \\ \hline
\textbf{RLC}         & 0.58         & 0.74          \\ \hline
\textbf{RPC}         & 0.78         & 0.79          \\ \hline
\textbf{ROD}         & 0.67         & 0.93          \\ \hline
\textbf{DIS}         & \_            & 0.99          \\ \hline
\textbf{Macro F1} & 0.73 & 0.88 \\ \hline
\end{tabular}
\caption{Label-wise Inter-Annotator agreement (F1 Scores). Dissent label instance absent in IT.}
\label{tab:interanno_pairwise_8labels}
\end{table}

\noindent \textbf{Adjudication and Data compilation:} 
Annotating RR is not a trivial task, and annotators can have disagreements. We followed a majority voting strategy over primary labels to determine the gold labels. There were a few cases ($\approx$ 5\%) where all the three legal experts assigned a different role to the same sentence. We asked the law professors to finalize the primary label in such cases. If the law professors decided to go with a label completely different from the three annotated labels, we went with their verdict. However, such cases were not frequent ($\approx$ 4\% of adjudicated cases). In this paper, for RR prediction, we concentrate on the primary labels and leave explorations of secondary and tertiary labels for future work. 


\begin{figure}[t]
     \centering
     \begin{subfigure}[b]{0.40\textwidth}
         \centering
         \includegraphics[width=\textwidth]{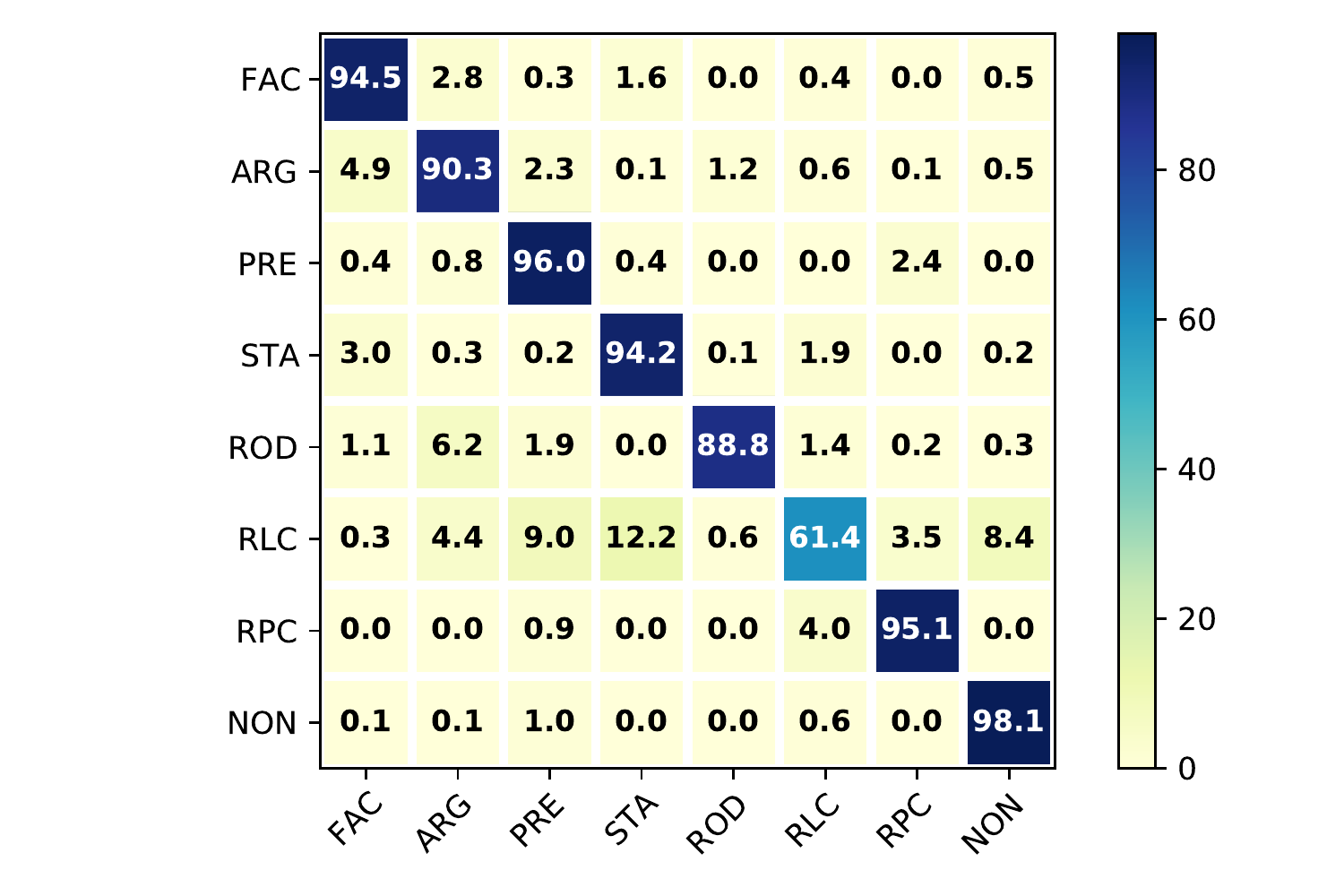}
         \caption{IT}
         \label{fig:IT_cm}
     \end{subfigure}
     \hfill
     \begin{subfigure}[b]{0.40\textwidth}
         \centering
         \includegraphics[width=\textwidth]{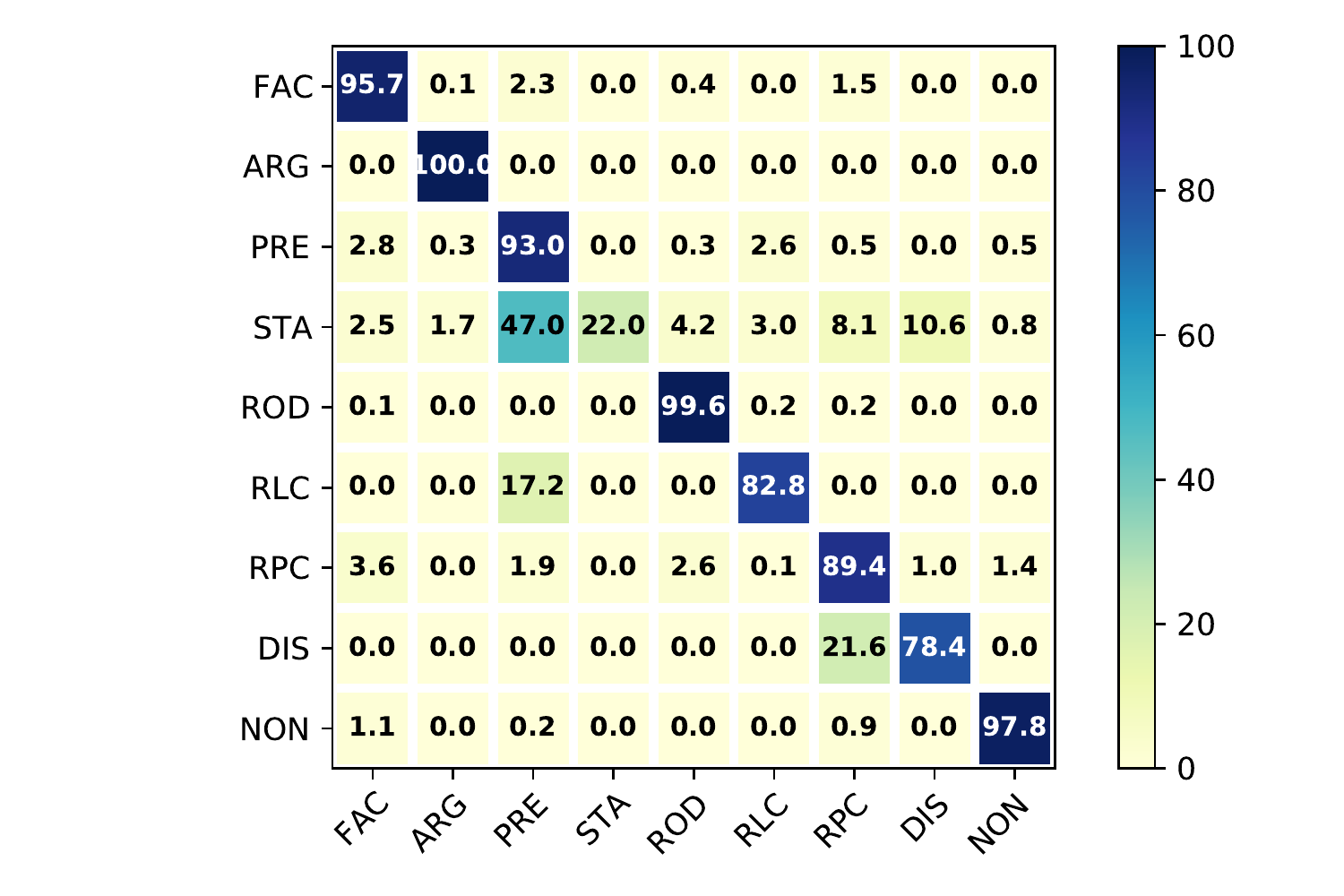}
         \caption{CL}
         \label{fig:cm}
     \end{subfigure}
     \hfill
        \caption{Confusion matrix between Annotators $A_{1}$ and $A_{3}$. Numbers represent \% agreement. Dissent label instance is absent in IT.}
        \label{fig:confusion-mat}
\end{figure}

\noindent \textbf{Inter-annotator Agreements:}
The Fleiss kappa \cite{fleiss2013statistical} between the annotators is 0.65 for the IT domain and 0.87 for the CL domain, indicating a substantial agreement between annotators. Additionally, as done in  \citet{bhattacharya2019identification} and \citet{malik-etal-2021-ildc}, we calculate the pair-wise inter-annotator F1 scores. To determine the agreement between the three annotators $A_{1}, A_{2}, A_{3}$ (each for IT and CL domain), we calculate the pairwise F1 scores (App. \ref{app:metrics}) between annotators $(A_{1},A_{2}), (A_{2},A_{3})$ and $(A_{3},A_{1})$. We average these pairwise scores for each label and further average them out. We report the label-wise F1 and Macro F1 in Table \ref{tab:interanno_pairwise_8labels}. The table shows that the agreements between domains differ (0.73 for IT vs. 0.88 for CL). This is mainly due to (as pointed by law professors) the presence of more precedents and a greater number of statutory provisions in IT laws. These factors combine to produce more subjectivity (relative to CL) when it comes to interpreting and retracing judicial decisions. The confusion matrix between the annotators $(A_{1},A_{3})$ is shown in Figure \ref{fig:confusion-mat} (more details in App. \ref{app:agreement}). 

\noindent \textbf{Analysis:} Annotation of judgments to identify RR is a challenging task even for legal experts. Several factors contribute to this challenge. Annotators need to glean and combine information non-trivially (e.g., facts and arguments presented, the implicit setting, and the context under which the events described in the case happened) to arrive at the label. Moreover, the annotator only has access to the current document, which is a secondary account of what actually happened in the court. These limitations certainly make the task of the annotator more difficult and leave them with no choice other than to make certain educated guesses when it comes to understanding the various nuances, both ostensible and probable, of certain RR. It should, however, be noted that such variation need not occur for every RR since not all the roles are equally susceptible to it. A cumulative effect of the aforementioned factors can be observed in the results of the annotation. The analysis provided by the three annotators in the case of CL bears close resemblance with each other. On the other hand, in the case of IT, the analysis provided by Users 1 and 3 bears a greater resemblance with each other, compared to the resemblance between Users 1 and 2, or between Users 2 and 3. On a different note, it is also observed that the rhetorical role where the annotators have differed between themselves the most has been the point of Ruling made by the Lower Court, followed by the Ratio. This also ties in with the argument that all rhetorical roles are not equally susceptible to the variation caused by the varying levels of success achieved by the different annotators in retracing the judicial thought pattern (details and case studies in App. \ref{app:annotation-analysis}).   

\section{Rhetorical Roles Prediction} \label{sec:models}



We would like to automate the process of segmenting a legal document, to develop ML models for the automation, we experiment with the task of Rhetorical Roles prediction. 

\noindent \textbf{Task Definition:} Given a legal document, $D$, containing the sentences $[s_{1}, s_{2}, ...s_{n}]$, the task of rhetorical role prediction is to predict the label (or role) $y_{i}$ for each sentence $s_{i} \in D$. 

\noindent \textbf{Baseline Models:} For the first set of baseline models, the task is modeled as a single sentence prediction task, where given the sentence  $s$, the model predicts the rhetorical role of the sentence. In this case, the context is ignored. We consider pre-trained BERT \cite{devlin-etal-2019-bert} and LEGAL-BERT \cite{chalkidis-etal-2020-legal} models for this. As another set of baseline models, we consider the task as a sequence labeling task, where the sequence of all the sentences in the document is given as input, and the model has to predict the RR label for each sentence. We used CRF with hand-crafted features \cite{bhattacharya2019identification} and BiLSTM network.

\begin{table}[t]
\small
\centering
\begin{tabular}{|c|c|c|}
\hline
\textbf{Model} & \textbf{Dataset} & \textbf{F1}  \\ \hline
SBERT-Shift    & IT               & 0.60        \\ \hline
SBERT-Shift    & CL               & 0.49               \\ \hline
SBERT-Shift    & IT+CL            & 0.47              \\ \hline
BERT-SC        & IT               & 0.66               \\ \hline
BERT-SC        & CL               & 0.64             \\ \hline
BERT-SC        & IT+CL            & 0.64              \\ \hline
\end{tabular}
\caption{Results for the auxillary task LSP} 
\label{tab:auxillary}
\end{table}

\noindent \textbf{Label Shift Prediction:} 
Rhetorical role labels do not change abruptly across sentences in a document, and the text tends to maintain topical coherence. Given the label $y$ for a sentence $s_{i}$ in the document, we hypothesize that the chances of shift (change) in the label for the next sentence $s_{i+1}$ are low. We manually verified this using the training set and observed that on average in a document, if the label of sentence $s_{i}$ is $y$, then 88\% of the times the label of the next sentence $s_{i+1}$ is same as $y$. Note that this is true only for consecutive sentences, but in general, label shift inertia fades as we try to predict beyond the second consecutive sentence. Since we are performing a sequence prediction task, this alone is not a good model for label prediction. Nevertheless, we think that this label shift inertia can provide a signal (via an auxiliary task) to the main sequence prediction model. Based on this observation, we define an auxiliary binary classification task: Label Shift Prediction (LSP), that aims to model the relationship between two sentences $s_{i}$ and $s_{i+1}$ and predict whether the labels $y_{i}$ for $s_{i}$ and $y_{i+1}$ for $s_{i+1}$ are different (shift occurs) or not. In particular, for each sentence pair $S = \{s_{i}, s_{i+1}\} \in D$, we define the label of LSP task, $Y=1$ if $y_{i}\neq y_{i+1}$, otherwise $Y=0$, here $y_{i}$ is the rhetorical role for sentence $s_{i}$.  
Note that for the full model at the inference time, the true label of a sentence is not provided; hence predicting a shift in label makes more sense than performing a binary prediction that the next sentence has the same label or not. We model the LSP task via two different models:


\noindent \textbf{SBERT-Shift:} We model the label shift via a Siamese network. In particular, we use the pre-trained SBERT model \cite{reimers2019sentence} to encode sentences $s_{i}$ and $s_{i+1}$ to get representations $e_{i}$ and $e_{i+1}$. The combination of these representations ($e_{i}\oplus e_{i+1}\oplus (e_{i}-e_{i+1})$) is passed through a feed-forward network to predict the shift. 

\noindent \textbf{BERT-SC:} We use the pre-trained BERT model and fine-tune it for the task of LSP. We model the input in the form of sentence semantic coherence task, $[CLS]\oplus s_{i}\oplus [SEP] \oplus s_{i+1} \oplus [SEP]$ to make the final prediction for shift. In general, the BERT-SC model performs better than SBERT-Shift (Table \ref{tab:auxillary}). 
Due to the superior performance of BERT-SC, we include it to provide label shift information to the final MTL model. The aim of our work is to predict RR, and we use label shift as auxiliary information even if it may not be predicted correctly at all times. As shown in results later, this limited information improves the performance.

\noindent \textbf{Proposed Models:} We propose two main models for the rhetorical role prediction: Label Shift Prediction based on BiLSTM-CRF and MTL models.

\begin{figure}[t]
\centering
\includegraphics[width=0.30\textwidth]{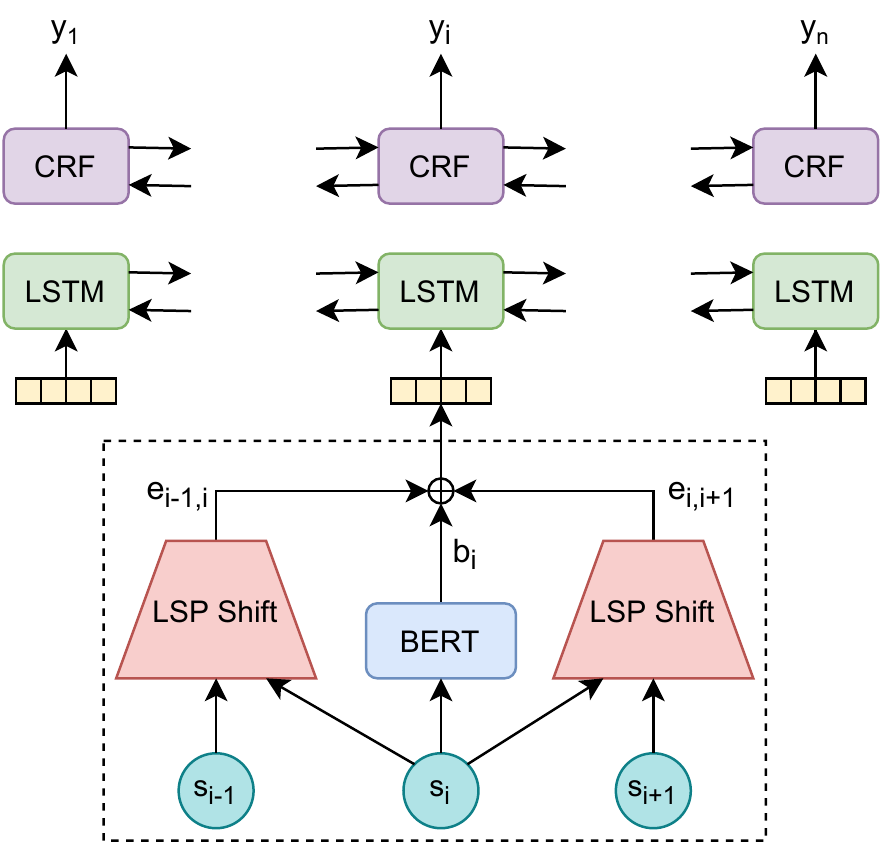}
\caption{LSP-BiLSTM-CRF Model}
\label{fig:auxRR}
\end{figure}

\noindent \textbf{LSP-BiLSTM-CRF:} Signal from label shift is used to aid the RR prediction in the LSP-BiLSTM-CRF model. The model consists of (Figure \ref{fig:auxRR}) a BiLSTM-CRF model with specialized input representation. Let the sentence embedding (from pre-trained BERT) corresponding to $i^{th}$ sentence be $b_{i}$. Let, the representation of the label shift (the layer before the softmax layer in LSP model) between current sentence and previous sentence pair $\{s_{i-1}, s_{i}\}$ be $e_{i-1,i}$. Similarly for the next pair ($\{s_{i}, s_{i+1}\}$) we get $e_{i,i+1}$. The sentence representation for $i^{th}$ sentence is given by $e_{i-1,i} \oplus b_{i} \oplus e_{i,i+1}$. This sentence representation goes as input to the BiLSTM-CRF model for RR prediction. 

\begin{figure}[t]
\centering
\includegraphics[width=0.40\textwidth]{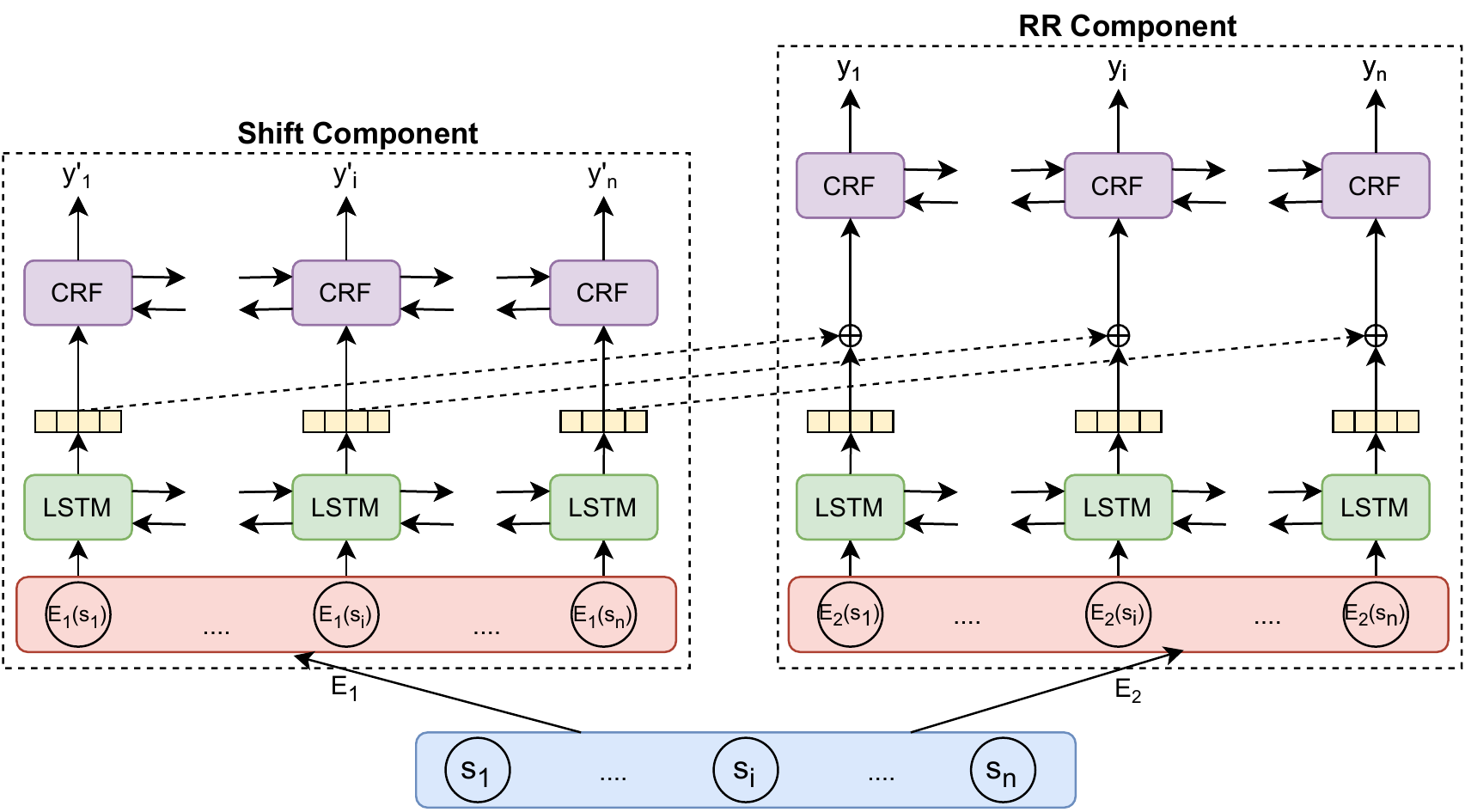}
\caption{MTL architecture for Rhetorical Role Labelling and Shift Prediction.}
\label{fig:MTLarch}
\vspace{-6mm}
\end{figure}

\noindent \textbf{MultiTask Learning (MTL):} We use the framework of Multitask learning, where rhetorical role prediction is the main task and label shift prediction is the auxiliary task. Sharing representations between the main and related tasks helps in better generalization on the main task \cite{multiTask-survey-2020}. The intuition is that a label shift would help the rhetorical role component make the correct prediction based on the prospective shift. The MTL model (Figure \ref{fig:MTLarch}) consists of two components: the shift detection component and the rhetorical role prediction component. The shift component predicts if a label shift occurs at $i^{th}$ position. The output of the BiLSTM layer of shift component is concatenated with the BiLSTM output of the rhetorical role component. The concatenated output is passed to a CRF layer for the final prediction of the rhetorical role. The loss for the model is given by: $L = \lambda L_{shift} + (1-\lambda ) L_{RR}$, where, $L_{shift}$ is the loss corresponding to label shift prediction and $L_{RR}$ is the loss corresponding to rhetorical role prediction, and hyperparameter $\lambda$ balances the importance of each of the task. If $\lambda$ is set to zero, we are back with our baseline BiLSTM-CRF model. Since there are two components, we experimented with sending the same encodings of sentences to both the components $(E_{1}=E_{2})$, as well as sending different encodings of the same sentence to both components $(E_{1} \neq E_{2})$. The proposed model is very different from the previously proposed BiLSTM-CRF by \citet{bhattacharya2019identification} that does not use any multitasking and label shift information.  

\section{Experiments, Results and Analysis} \label{sec:experiments}

\begin{figure}[t]
    \centering
    \includegraphics[width=0.30\textwidth]{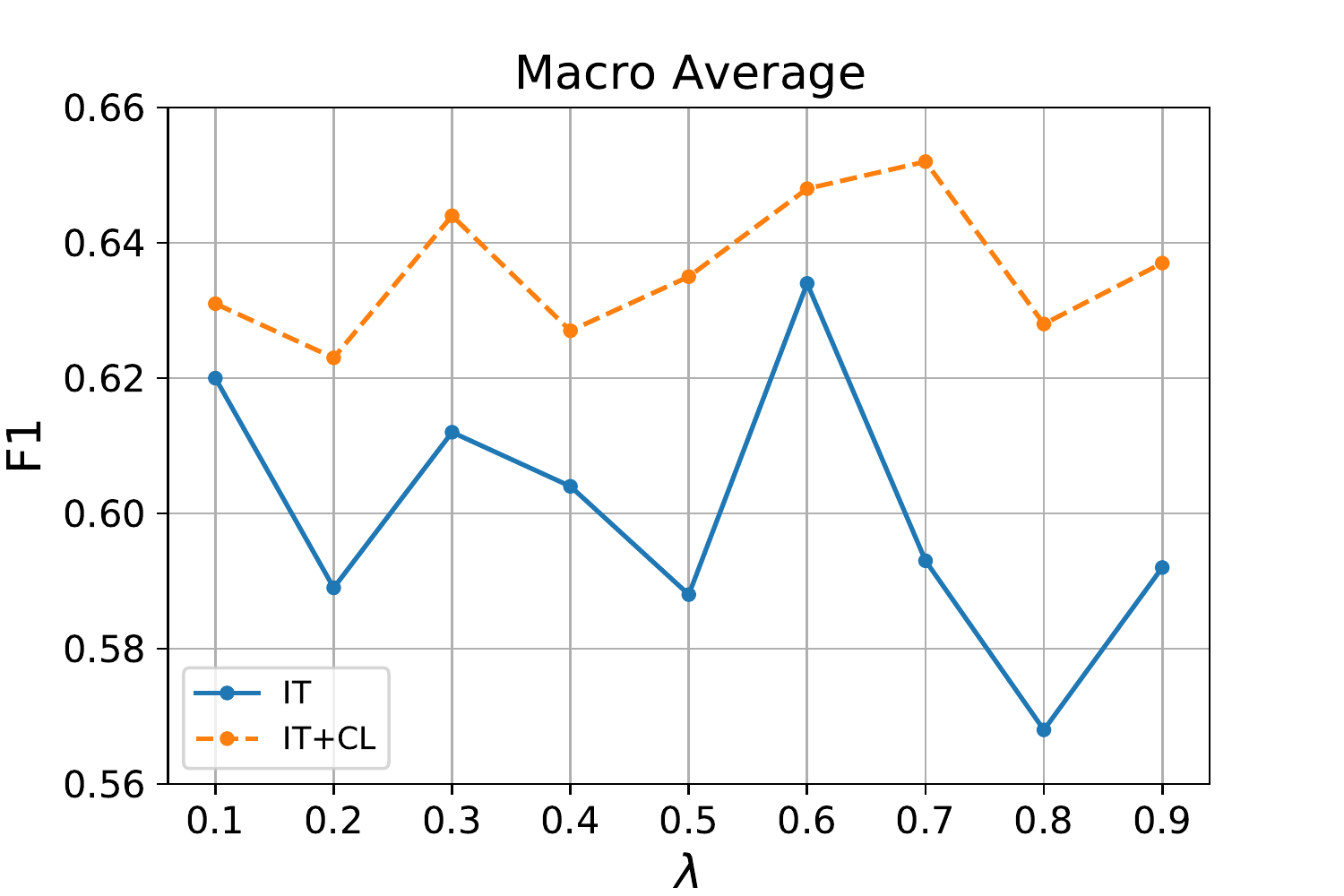}
    \caption{Variation of F1 score with $\lambda$ on IT and IT+CL domain}
    \label{app:fig:lambdavariation}
\end{figure}

Due to the complexity of the task of RR prediction and to be comparable with the existing baseline systems, for experiments, we consider 7 main labels (FAC, ARG, PRE, ROD, RPC, RLC, and STA). We plan to explore all fine-grained RR label (13) predictions in the future. Based on recommendations by legal experts, we ignore sentences with NON (None) label (about 4\% for IT and 0.5\% for CL) (more details in App. \ref{app:reducedLabels}). Further, the IT domain did not have any instance of dissent (DIS) label, and CL has only three documents with very few DIS instances. Based on consultations with law experts, we discarded DIS sentences (more details in App. \ref{app:reducedLabels}).  
We randomly split (at document level) IT/CL into 80\% train, 10\% validation, and 10\% test set. In contrast to \citet{bhattacharya2019identification}, we did not perform cross-validation for better comparison across different models. 
We also experiment with a combined dataset of IT and CL (IT+CL); the splits are made by combining individual train/val/test split of IT and CL. We experimented with a number of baseline models (Table \ref{tab:RRresultsITCL}, \ref{tab:rrresultscombineddomain}). In particular, we considered BiLSTM with sent2vec embeddings \cite{bhattacharya2019identification}, non-contextual models (single sentence) like BERT \cite{devlin-etal-2019-bert}, LegalBERT \cite{chalkidis-etal-2020-legal} and BERT-neighbour (we take both left and right neighboring sentences in addition to the sentence of interest). We also considered sentence-level sequence prediction models (contextual models): CRF model using handcrafted features provided by \citet{bhattacharya2019identification}, different variants of BiLSTM-CRF, one with handcrafted features, with sent2vec embeddings, with BERT embeddings, and with MLM embeddings. We finetuned BERT with Masked Language Modeling (MLM) objective on the train set to obtain MLM embeddings (CLS embedding) for each of the sentences (App. \ref{app:model} has hyperparameters, training schedule, and compute settings). We use the Macro F1 metric for evaluation (App. \ref{app:metrics}). We tuned the hyperparameter $\lambda$ of the MTL loss function using the validation set. We trained the MTL model with $\lambda \in  [0.1,0.9]$ with strides of $0.1$ (Figure \ref{app:fig:lambdavariation}). $\lambda=0.6$ performs the best for the IT domain and performs competitively on the combined domains.

\begin{table}[t]\small
\centering
\begin{tabular}{|l|c|c|}
\hline
\textbf{Model}                    & \textbf{IT (F1)} & \textbf{CL (F1)} \\ \hline
BERT                              & 0.56             & 0.52             \\ \hline
BERT-neighbor                     & 0.53            & 0.51                \\ \hline
LEGAL-BERT                        & 0.55            & 0.53             \\ \hline
CRF (Handcrafted)        & 0.55                & 0.52                \\ \hline
BiLSTM (sent2vec)                 & 0.55                & 0.54                \\ \hline
BiLSTM-CRF (handcraft) & 0.57                & 0.56                \\ \hline
BiLSTM-CRF (sent2vec)             & 0.59            & 0.61                \\ \hline
BiLSTM-CRF (BERT emb)             & 0.63            & 0.63                \\ \hline
BiLSTM-CRF (MLM emb)         & 0.58          & 0.60                \\ \hline
\textit{LSP (SBERT)}      & \textit{0.64}            & \textit{0.63}                \\ \hline
\textit{LSP (BERT-SC)} $\bullet$          & \textit{0.65}            & \textit{0.68}                \\ \hline
\textit{MTL (MLM emb)}         & \textit{0.67 }           & \textit{0.67}                \\ \hline
\textit{MTL (BERT-SC)} $\star$ $\diamond$       & \textit{\textbf{0.70}$\pm${0.02}}    & \textit{\textbf{0.69}$\pm$0.01}       \\ \hline
\end{tabular}
\caption{Results of baseline and proposed models on IT and CL. LSP and MTL refer to the LSP-BiLSTM-CRF and MTL-BiLSTM-CRF models respectively. $\bullet$ LSP result is significant with $p \le 0.05$ in comparison to baseline (BiLSTM-CRF(sent2vec)). Similarly, MTL (BERT-SC) has significant result in comparison to baseline ($\diamond$, $p \le 0.05$). MTL (BERT-SC) is significant w.r.t. LSP ($\star$, $p \le 0.05$).} 
\label{tab:RRresultsITCL}
\vspace{-2mm}
\end{table}

\begin{table}[t]\small
\centering
\begin{tabular}{|l|c|}
\hline
\textbf{Model}           & \textbf{IT+CL (F1)} \\ \hline
BiLSTM-CRF (sent2vec)    & 0.65                   \\ \hline
BiLSTM-CRF (BERT embs)   & 0.63                   \\ \hline
LSP-BiLSTM-CRF (BERT-SC) & 0.67                   \\ \hline
MTL-BiLSTM-CRF (BERT-SC)          & \textbf{0.70}$\pm$0.01          \\ \hline
\end{tabular}
\caption{Results of baseline and proposed models on combined dataset (IT+CL)}
\label{tab:rrresultscombineddomain}
\vspace{-4mm}
\end{table}

\begin{table}[t]
\small
\centering
\begin{tabular}{|l|l|l|}
\hline
\textbf{Label}       & \textbf{IT}   & \textbf{CL}    \\ \hline
\textbf{AR}          & 0.67$\pm$0.010         & 0.78$\pm$0.005          \\ \hline
\textbf{FAC}         & 0.78$\pm$0.020         & 0.75$\pm$0.010         \\ \hline
\textbf{PR}          & 0.69$\pm$0.005           & 0.62$\pm$0.005           \\ \hline
\textbf{STA}         & 0.79$\pm$0.020        & 0.82$\pm$0.020          \\ \hline
\textbf{RLC}         & 0.62$\pm$0.005         & 0.53$\pm$0.005          \\ \hline
\textbf{RPC}         & 0.70$\pm$0.010         & 0.71$\pm$0.010          \\ \hline
\textbf{ROD}         & 0.66$\pm$0.005         & 0.65$\pm$0.005          \\ \hline 
\textbf{Macro F1} & 0.70$\pm$0.020 & 0.69$\pm$0.010 \\ \hline
\end{tabular}
\caption{Label-wise average (across 6 runs) F1 scores of MTL-BiLSTM-CRF (BERT-SC) model.}
\vspace{-5mm}
\label{tab:predictedlabelwise}
\end{table}


\noindent \textbf{Results and Analysis:} Among the baseline models (Table \ref{tab:RRresultsITCL}), we note that LEGAL-BERT performs slightly better on the CL domain but slightly worse on the IT domain when compared to pre-trained BERT. It might be attributed to that LEGAL-BERT (trained on EU legal documents, which also has European competition law) is not trained on Indian IT law documents. Using BERT embeddings with BiLSTM-CRF provides better results. Both the proposed approaches outperform the previous approaches by a substantial margin. The MTL approach (with $\lambda=0.6$) provides the best results on both datasets with an average (over six runs) F1 score of $0.70$ (standard deviation of $0.02$) on the IT domain,  an average F1 of $0.69 (\pm 0.01)$ on CL domain, and an average F1 score of $0.71 (\pm 0.01)$ for the combined domain. The MTL model shows variance across runs; hence we average the results. Other models were reasonably stable across runs. 

We use the LSP shift component with BERT-SC as the encoder $E_{1}$ and the pre-trained BERT model as the encoder $E_{2}$ in our MTL architecture. We did not use SBERT since it was under-performing when compared to BERT-SC. We provide the label-wise F1 scores for the MTL model in Table \ref{tab:predictedlabelwise}. Note the high performance on the FAC label and low performance on the RLC label; this is similar to what we observe for annotators (Table \ref{tab:interanno_pairwise_8labels}). Also, the MTL model performs better on the AR label in the CL domain than the IT domain. An opposite trend can be observed for the RLC label. The contribution of the LSP task is evident from the superior performance. We conduct the ablation study of our MTL architecture from multiple aspects. Instead of using shift embeddings from BERT-SC as the encoder $E_{1}$, we use a BERT model fine-tuned upon the MLM task on the IT and CL domain. However, we obtain a comparatively lower score (see App. \ref{app:model}). This observation yet again points towards the significance of the LSP in the task of rhetorical role prediction (results on other encoders in App. \ref{app:model}). The results have two interesting observations: firstly, MTL model performance on IT cases comes close to the average inter-annotator agreement. In the case of CL, there is a gap. Secondly, for the model, the performance on the IT domain is better than the CL domain, but in the case of annotators opposite trend was observed. We do not know the exact reason for this, but the legal experts pointed out that this is possible because the selected documents might be restricted to specific sections of the IT law and model learned solely from these documents alone without any other external knowledge. However, annotators, having knowledge of the entire IT law,  might have looked from a broader perspective.

\begin{table}[t]
\small
\centering
\begin{tabular}{|c|c|c|c|}
\hline
\textbf{\begin{tabular}[c]{@{}c@{}}Train\\ Dataset\end{tabular}} & \textbf{\begin{tabular}[c]{@{}c@{}}Test\\ Dataset\end{tabular}} & \textbf{\begin{tabular}[c]{@{}c@{}}BiLSTM-CRF\\ (sent2vec)\end{tabular}} & \textbf{MTL}                                             \\ \hline
$G$\textsubscript{train}                                                         & $G$\textsubscript{test}                                                         & \begin{tabular}[c]{@{}c@{}}0.55\\ \end{tabular}                      & \begin{tabular}[c]{@{}c@{}}\textbf{0.59}\\ \end{tabular}      \\ \hline
$G$\textsubscript{train}                                                         & $CL$\textsubscript{test}                                                        & \begin{tabular}[c]{@{}c@{}}0.48\\ (12.78\%)\end{tabular}                 & \begin{tabular}[c]{@{}c@{}}\textbf{0.50}\\ (15.25\%)\end{tabular} \\ \hline
$G$\textsubscript{train}                                                         & $IT$\textsubscript{test}                                                        & \begin{tabular}[c]{@{}c@{}}0.41\\ (25.45\%)\end{tabular}                 & \begin{tabular}[c]{@{}c@{}}\textbf{0.46}\\ (22.03\%)\end{tabular} \\ \hline
$G$\textsubscript{train}                                                         & (IT+CL)\textsubscript{test}                                                     & \begin{tabular}[c]{@{}c@{}}0.42\\ (23.64\%)\end{tabular}                 & \begin{tabular}[c]{@{}c@{}}\textbf{0.48}\\ (18.64\%)\end{tabular} \\ \hline
(IT+CL)\textsubscript{train}                                                     & $G$\textsubscript{test}                                                         & \begin{tabular}[c]{@{}c@{}}0.60\\ \end{tabular}                      & \begin{tabular}[c]{@{}c@{}}\textbf{0.63}\\ \end{tabular}      \\ \hline
\end{tabular}
\caption{Domain transfer experiments to compare the performance of MTL-BiLSTM-CRF with the baseline BiLSTM-CRF. The number in parenthesis denotes $\Delta_{G}$ : the \% difference between the performance on $G$\textsubscript{test} and the new domain.} 
\label{tab:domaintransfer}
\vspace{-6mm}
\end{table}

\noindent \textbf{Domain Transfer:} In order to check the generalization capabilities of the MTL model compared to the baseline model, we conducted some domain transfer experiments. We experimented with a RR dataset of $50$ documents (referred to as $G$) by \citet{bhattacharya2019identification}. $G$ dataset comes from a different legal sub-domain (criminal and civil cases) with very less overlap with IT and CL. We tried different combinations of train and test datasets of IT, CL, and $G$. Note that $G$ (criminal and civil cases) has very less overlap with IT and CL cases, so practically, it is a different domain. The results are in Table \ref{tab:domaintransfer}. 
We can observe that the MTL model generalizes better across the domains than the baseline model. Both the models perform better on the $G$\textsubscript{test} when the combined (IT+CL)\textsubscript{train} set is used. This points towards better generalization. 

\noindent \textbf{Model distillation:} RR annotation is a tedious process, however, there is an abundance of unlabelled legal documents. We experimented with semi-supervised techniques to leverage the unlabelled data. In particular, we tried a self-training based approach \cite{xie2020self}. The idea is to learn a teacher model $\theta_{tea}$ on the labelled data $D_{L}$. The teacher model is then used to generate hard labels on unlabeled sentences $s_{u} \in d_{i}$: $\hat{y}_{i} = f_{\theta_{tea}}(\hat{d_{i}})\ \forall \hat{d_{i}} \in D_{U}$. Next, a student model $\theta_{stu}$ is learned on labeled and unlabeled sentences, with the loss function for student training given by: $ L_{ST} = \frac{1}{|D_{L}|}\sum_{d_{j}\in D_{L}}L( f_{\theta_{stu}}(d_{j}),y_{j}) + \frac{\alpha_{U}}{D_{U}}\sum_{ \hat{d_{i}}\in D_{U} }L(f_{\theta_{stu}}(\hat{d}_{i}), \hat{y_{i}}) $. Here, $\alpha_{U}$ is a weighing hyperparameter between the labelled and unlabelled data (details in App. \ref{app:model}). The process can be iterated and the final distilled model is used for prediction. The results of model distillation are shown in Table \ref{tab:model-distillation} for two iterations (initializing the teacher model of the current iteration as the learned student model of the previous iteration; further iterations do not improve results). MTL model was run just once, due to variance it shows F1 of 0.68. The results improve for majority of labels with an increment of 0.11 F1 score for the RLC label in the first iteration. Also, the variance of F1 scores across labels decreases. 

\vspace{-2mm}
\subsection{Application of Rhetorical Role to Judgment Prediction} \label{app:judgment} \vspace{-2mm}

To check the applicability of RR in downstream applications, as a use-case, we experimented with how RR could contribute towards judgment prediction (ethical concerns discussed later). We use the legal judgment corpus (ILDC) provided by \citet{malik-etal-2021-ildc} and fine-tune a pre-trained BERT model on the train set of ILDC for the task of judgment prediction on the last 512 tokens of the documents.  \citet{malik-etal-2021-ildc} observed that training on the last 512 (also the max size of the input to BERT) tokens of a legal document give the best results; we use the same setting. We use this trained model directly for predicting the outcome on 84 IT/CL cases. We removed text corresponding to the final decisions (and extracted gold decisions) from these documents with the help of legal experts. In the first experiment, we use the last 512 tokens of IT/CL cases for prediction. To study the effect of RRs, in another experiment, we extract the sentences corresponding to gold ratio (ROD) and ruling (RPC) RR labels in IT/CL documents and use this as input to the BERT model. We consider these two RR only since, by definition, these sentences denote the principles and the decision of the court related to the issues in the proceedings. There were no ROD or RPC labels for some documents (16 out of 100 for both IT and CL); we removed these in both experiments. The results are shown in Table \ref{tab:JP}. Using the gold RR gives a boost to the F1 score. 
We also experimented with using predicted RR, and the performance was comparable to that of the BERT model. 

To explore how predicted rhetorical roles would perform on judgment prediction task, we perform the following experiment. We use our best performing model MTL (BERT-SC), trained on the combined IT+CL domain to check the applicability of rhetorical roles for the task of Judgment Prediction. In the first step, we obtain the predicted rhetorical roles for each sentence in the documents. Next, we select the sentences labeled as ROD or RPC\footnote{We select only these two labels since by definition, these sentences provide the necessary cues towards the judgment.}. Third, we use a BERT base model fine-tuned on the last 512 tokens of each document in the ILDC corpus \cite{malik-etal-2021-ildc} and use it to predict the judgment of the test set documents, given only the predicted ROD and RPC sentences. We compare the results by the MTL model and BiLSTM-CRF baseline on performing judgment prediction with predicted rhetorical roles. Refer to Appendix Table \ref{app:tab:jp} for the results. Since RR prediction for ROD and RPC is not perfect, improving it would greatly enhance the results as shown in Table \ref{tab:JP}. 

\begin{table}[t]
\small
\centering
\begin{tabular}{|l|l|c|}
\hline
\textbf{Model} & \textbf{IT+CL docs} & \textbf{F1} \\ \hline
BERT-ILDC      & last 512 tokens         & 0.55           \\ \hline
BERT-ILDC      & Gold ROD \& RPC         & \textbf{0.58}           \\ \hline
\end{tabular}
\caption{Judgment prediction using RR. The model using gold ROD and RPC is found to be statistically significant ($p \le 0.05$).} 
\label{tab:JP}
\vspace{-3mm}
\end{table}

\begin{table}[t]
\small
\centering
\begin{tabular}{|l|c|c|c|}
\hline
\textbf{Label}       & \textbf{Base MTL}   & \textbf{Dist. Iter 1} & \textbf{Dist. Iter 2}    \\ \hline
\textbf{AR}          & 0.62         & 0.70     & \textbf{0.70}     \\ \hline
\textbf{FAC}         & 0.74         & \textbf{0.75}   & 0.73       \\ \hline
\textbf{PR}          & 0.68           & 0.72    & \textbf{0.74}       \\ \hline
\textbf{STA}         & 0.76        & \textbf{0.77}      & 0.75   \\ \hline
\textbf{RLC}         & 0.59         & \textbf{0.70}   & 0.70       \\ \hline
\textbf{RPC}         & 0.67         & 0.63      & \textbf{0.73}    \\ \hline
\textbf{ROD}         & 0.68         & 0.66     & \textbf{0.68}     \\ \hline
\textbf{Macro F1} & 0.68 & 0.71 & \textbf{0.72}\\ \hline
\end{tabular}
\caption{Model Distillation: F1 scores of MTL-BiLSTM-CRF (BERT-SC) model after two distillation iterations on the IT domain.}
\label{tab:model-distillation}
\vspace{-6mm}
\end{table}

\section{Conclusion}
\vspace{-2mm}
We introduce a new corpus annotated with rhetorical roles. We proposed a new MTL model that uses label shift information for predicting labels. We further showed via domain transfer experiments the generalizability of the model. Since RR are tedious to annotate, we showed the possibility of using model distillation techniques to improve the system. In the future, we plan to explore cross-domain transfer techniques to perform RR identification in legal documents in other Indian languages. Nevertheless, we plan to grow the corpus. We also plan to apply RR models for other legal tasks such as summarization and information extraction. 

\section*{Acknowledgements}
\vspace{-2mm}
We would like to thank anonymous reviewers for their insightful comments. We would like to thank student research assistants Tridib Mandal, Chirag Mittal, Shefali Deshmukh, Shailja Beria, and Syamantak Sinha from West Bengal National University of Juridical Sciences (WBNUJS) for annotating the documents. This work would not have been possible without their help.


\bibliography{references}

\begin{thebibliography}{35}
\expandafter\ifx\csname natexlab\endcsname\relax\def\natexlab#1{#1}\fi

\bibitem[{Bhattacharya et~al.(2019)Bhattacharya, Paul, Ghosh, Ghosh, and
  Wyner}]{bhattacharya2019identification}
Paheli Bhattacharya, Shounak Paul, Kripabandhu Ghosh, Saptarshi Ghosh, and Adam
  Wyner. 2019.
\newblock \href {http://arxiv.org/abs/1911.05405} {Identification of rhetorical
  roles of sentences in indian legal judgments}.
\newblock \emph{CoRR}, abs/1911.05405.

\bibitem[{Chalkidis et~al.(2019)Chalkidis, Androutsopoulos, and
  Aletras}]{chalkidis-etal-2019-neural}
Ilias Chalkidis, Ion Androutsopoulos, and Nikolaos Aletras. 2019.
\newblock \href {https://doi.org/10.18653/v1/P19-1424} {Neural legal judgment
  prediction in {E}nglish}.
\newblock In \emph{Proceedings of the 57th Annual Meeting of the Association
  for Computational Linguistics}, pages 4317--4323, Florence, Italy.
  Association for Computational Linguistics.

\bibitem[{Chalkidis et~al.(2020)Chalkidis, Fergadiotis, Malakasiotis, Aletras,
  and Androutsopoulos}]{chalkidis-etal-2020-legal}
Ilias Chalkidis, Manos Fergadiotis, Prodromos Malakasiotis, Nikolaos Aletras,
  and Ion Androutsopoulos. 2020.
\newblock \href {https://doi.org/10.18653/v1/2020.findings-emnlp.261}
  {{LEGAL}-{BERT}: The muppets straight out of law school}.
\newblock In \emph{Findings of the Association for Computational Linguistics:
  EMNLP 2020}, pages 2898--2904, Online. Association for Computational
  Linguistics.

\bibitem[{Chen et~al.(2019)Chen, Cai, Dai, Dai, and
  Ding}]{chen-etal-2019-charge}
Huajie Chen, Deng Cai, Wei Dai, Zehui Dai, and Yadong Ding. 2019.
\newblock \href {https://doi.org/10.18653/v1/D19-1667} {Charge-based prison
  term prediction with deep gating network}.
\newblock In \emph{Proceedings of the 2019 Conference on Empirical Methods in
  Natural Language Processing and the 9th International Joint Conference on
  Natural Language Processing (EMNLP-IJCNLP)}, pages 6362--6367, Hong Kong,
  China. Association for Computational Linguistics.

\bibitem[{Crawshaw(2020)}]{multiTask-survey-2020}
Michael Crawshaw. 2020.
\newblock \href {http://arxiv.org/abs/2009.09796} {Multi-task learning with
  deep neural networks: {A} survey}.
\newblock \emph{CoRR}, abs/2009.09796.

\bibitem[{de~Castilho et~al.(2016)de~Castilho, Mujdricza-Maydt, Yimam,
  Hartmann, Gurevych, Frank, and Biemann}]{de2016web}
Richard~Eckart de~Castilho, Eva Mujdricza-Maydt, Seid~Muhie Yimam, Silvana
  Hartmann, Iryna Gurevych, Anette Frank, and Chris Biemann. 2016.
\newblock A web-based tool for the integrated annotation of semantic and
  syntactic structures.
\newblock In \emph{Proceedings of the Workshop on Language Technology Resources
  and Tools for Digital Humanities (LT4DH)}, pages 76--84.

\bibitem[{Devlin et~al.(2019)Devlin, Chang, Lee, and
  Toutanova}]{devlin-etal-2019-bert}
Jacob Devlin, Ming-Wei Chang, Kenton Lee, and Kristina Toutanova. 2019.
\newblock {BERT}: {P}re-training of {D}eep {B}idirectional {T}ransformers for
  {L}anguage {U}nderstanding.
\newblock In \emph{Proceedings of the 2019 Conference of the North {A}merican
  Chapter of the Association for Computational Linguistics: Human Language
  Technologies, Volume 1 (Long and Short Papers)}, pages 4171--4186,
  Minneapolis, Minnesota. Association for Computational Linguistics.

\bibitem[{Fleiss et~al.(2013)Fleiss, Levin, and Paik}]{fleiss2013statistical}
Joseph~L Fleiss, Bruce Levin, and Myunghee~Cho Paik. 2013.
\newblock \emph{Statistical methods for rates and proportions}.
\newblock john wiley \& sons.

\bibitem[{Hu et~al.(2018)Hu, Li, Tu, Liu, and Sun}]{hu-etal-2018-shot}
Zikun Hu, Xiang Li, Cunchao Tu, Zhiyuan Liu, and Maosong Sun. 2018.
\newblock \href {https://aclanthology.org/C18-1041} {Few-shot charge prediction
  with discriminative legal attributes}.
\newblock In \emph{Proceedings of the 27th International Conference on
  Computational Linguistics}, pages 487--498, Santa Fe, New Mexico, USA.
  Association for Computational Linguistics.

\bibitem[{Jackson et~al.(2003)Jackson, Al-Kofahi, Tyrrell, and
  Vachher}]{jackson2003information}
Peter Jackson, Khalid Al-Kofahi, Alex Tyrrell, and Arun Vachher. 2003.
\newblock Information extraction from case law and retrieval of prior cases.
\newblock \emph{Artificial Intelligence}, 150(1-2):239--290.

\bibitem[{Jiang et~al.(2018)Jiang, Ye, Luo, Chao, and
  Ma}]{jiang-etal-2018-interpretable}
Xin Jiang, Hai Ye, Zhunchen Luo, WenHan Chao, and Wenjia Ma. 2018.
\newblock \href {https://aclanthology.org/C18-2032} {Interpretable rationale
  augmented charge prediction system}.
\newblock In \emph{Proceedings of the 27th International Conference on
  Computational Linguistics: System Demonstrations}, pages 146--151, Santa Fe,
  New Mexico. Association for Computational Linguistics.

\bibitem[{Kalamkar et~al.(2022{\natexlab{a}})Kalamkar, Tiwari, Agarwal, Karn,
  Gupta, Raghavan, and Modi}]{kalamkar-EtAl:2022:LREC}
Prathamesh Kalamkar, Aman Tiwari, Astha Agarwal, Saurabh Karn, Smita Gupta,
  Vivek Raghavan, and Ashutosh Modi. 2022{\natexlab{a}}.
\newblock Corpus for automatic structuring of legal documents.
\newblock In \emph{Proceedings of the Language Resources and Evaluation
  Conference}, pages 4420--4429, Marseille, France. European Language Resources
  Association.

\bibitem[{Kalamkar et~al.(2022{\natexlab{b}})Kalamkar, Tiwari, Agarwal, Karn,
  Gupta, Raghavan, and Modi}]{kalamkar-etal-2022-corpus}
Prathamesh Kalamkar, Aman Tiwari, Astha Agarwal, Saurabh Karn, Smita Gupta,
  Vivek Raghavan, and Ashutosh Modi. 2022{\natexlab{b}}.
\newblock \href {https://aclanthology.org/2022.lrec-1.470} {Corpus for
  automatic structuring of legal documents}.
\newblock In \emph{Proceedings of the Thirteenth Language Resources and
  Evaluation Conference}, pages 4420--4429, Marseille, France. European
  Language Resources Association.

\bibitem[{Kapoor et~al.(2022)Kapoor, Dhawan, Goel, T~H, Bhatnagar, Agrawal,
  Agrawal, Bhattacharya, Kumaraguru, and Modi}]{kapoor-etal-2022-hldc}
Arnav Kapoor, Mudit Dhawan, Anmol Goel, Arjun T~H, Akshala Bhatnagar, Vibhu
  Agrawal, Amul Agrawal, Arnab Bhattacharya, Ponnurangam Kumaraguru, and
  Ashutosh Modi. 2022.
\newblock \href {https://doi.org/10.18653/v1/2022.findings-acl.278} {{HLDC}:
  {H}indi legal documents corpus}.
\newblock In \emph{Findings of the Association for Computational Linguistics:
  ACL 2022}, pages 3521--3536, Dublin, Ireland. Association for Computational
  Linguistics.

\bibitem[{Katju(2019)}]{backlogcases2019}
Justice~Markandey Katju. 2019.
\newblock Backlog of cases crippling judiciary.
\newblock \url{https://perma.cc/D8V4-L566}.

\bibitem[{Lagos et~al.(2010)Lagos, Segond, Castellani, and
  O’Neill}]{lagos2010event}
Nikolaos Lagos, Frederique Segond, Stefania Castellani, and Jacki O’Neill.
  2010.
\newblock Event extraction for legal case building and reasoning.
\newblock In \emph{International Conference on Intelligent Information
  Processing}, pages 92--101. Springer.

\bibitem[{Leitner et~al.(2019)Leitner, Rehm, and
  Moreno-Schneider}]{10.1007/978-3-030-33220-4_20}
Elena Leitner, Georg Rehm, and Julian Moreno-Schneider. 2019.
\newblock Fine-grained named entity recognition in legal documents.
\newblock In \emph{Semantic Systems. The Power of AI and Knowledge Graphs},
  pages 272--287, Cham. Springer International Publishing.

\bibitem[{Malik et~al.(2021)Malik, Sanjay, Nigam, Ghosh, Guha, Bhattacharya,
  and Modi}]{malik-etal-2021-ildc}
Vijit Malik, Rishabh Sanjay, Shubham~Kumar Nigam, Kripabandhu Ghosh,
  Shouvik~Kumar Guha, Arnab Bhattacharya, and Ashutosh Modi. 2021.
\newblock \href {https://doi.org/10.18653/v1/2021.acl-long.313} {{ILDC} for
  {CJPE}: {I}ndian legal documents corpus for court judgment prediction and
  explanation}.
\newblock In \emph{Proceedings of the 59th Annual Meeting of the Association
  for Computational Linguistics and the 11th International Joint Conference on
  Natural Language Processing (Volume 1: Long Papers)}, pages 4046--4062,
  Online. Association for Computational Linguistics.

\bibitem[{Moens et~al.(1999)Moens, Uyttendaele, and
  Dumortier}]{moens1999abstracting}
Marie-Francine Moens, Caroline Uyttendaele, and Jos Dumortier. 1999.
\newblock Abstracting of legal cases: the potential of clustering based on the
  selection of representative objects.
\newblock \emph{Journal of the American Society for Information Science},
  50(2):151--161.

\bibitem[{{National Judicial Data Grid}(2021)}]{njdc-district}
{National Judicial Data Grid}. 2021.
\newblock National judicial data grid statistics.
\newblock \url{https://www.njdg.ecourts.gov.in/njdgnew/index.php}.

\bibitem[{Nejadgholi et~al.(2017)Nejadgholi, Bougueng, and
  Witherspoon}]{nejadgholi2017semi}
Isar Nejadgholi, Renaud Bougueng, and Samuel Witherspoon. 2017.
\newblock A semi-supervised training method for semantic search of legal facts
  in canadian immigration cases.
\newblock In \emph{JURIX}, pages 125--134.

\bibitem[{Reimers and Gurevych(2019)}]{reimers2019sentence}
Nils Reimers and Iryna Gurevych. 2019.
\newblock Sentence-bert: Sentence embeddings using siamese bert-networks.
\newblock \emph{arXiv preprint arXiv:1908.10084}.

\bibitem[{Saravanan et~al.(2008)Saravanan, Ravindran, and
  Raman}]{saravanan2008automatic}
M~Saravanan, Balaraman Ravindran, and S~Raman. 2008.
\newblock Automatic identification of rhetorical roles using conditional random
  fields for legal document summarization.
\newblock In \emph{Proceedings of the Third International Joint Conference on
  Natural Language Processing: Volume-I}.

\bibitem[{Savelka and Ashley(2018)}]{savelka2018segmenting}
Jaromir Savelka and Kevin~D Ashley. 2018.
\newblock Segmenting us court decisions into functional and issue specific
  parts.
\newblock In \emph{JURIX}, pages 111--120.

\bibitem[{Skylaki et~al.(2021)Skylaki, Oskooei, Bari, Herger, and
  Kriegman}]{9679940}
Stavroula Skylaki, Ali Oskooei, Omar Bari, Nadja Herger, and Zac Kriegman.
  2021.
\newblock \href {https://doi.org/10.1109/ICDMW53433.2021.00086} {Legal entity
  extraction using a pointer generator network}.
\newblock In \emph{2021 International Conference on Data Mining Workshops
  (ICDMW)}, pages 653--658.

\bibitem[{Strickson and De~La~Iglesia(2020)}]{strickson2020legal}
Benjamin Strickson and Beatriz De~La~Iglesia. 2020.
\newblock Legal judgement prediction for uk courts.
\newblock In \emph{Proceedings of the 2020 The 3rd International Conference on
  Information Science and System}, pages 204--209.

\bibitem[{Taxmann(2021)}]{taxmann2021}
Taxmann. 2021.
\newblock Interpretation of statutes: Strict versus liberal construction.
\newblock \url{https://tinyurl.com/2p85h3xd}.

\bibitem[{Tran et~al.(2019)Tran, Nguyen, and Satoh}]{tran2019building}
Vu~Tran, Minh~Le Nguyen, and Ken Satoh. 2019.
\newblock Building legal case retrieval systems with lexical matching and
  summarization using a pre-trained phrase scoring model.
\newblock In \emph{Proceedings of the Seventeenth International Conference on
  Artificial Intelligence and Law}, pages 275--282.

\bibitem[{Venturi(2012)}]{venturi2012design}
Giulia Venturi. 2012.
\newblock Design and development of temis: a syntactically and semantically
  annotated corpus of italian legislative texts.
\newblock In \emph{Proceedings of the Workshop on Semantic Processing of Legal
  Texts (SPLeT 2012)}, pages 1--12.

\bibitem[{Walker et~al.(2019)Walker, Pillaipakkamnatt, Davidson, Linares, and
  Pesce}]{walker2019automatic}
Vern~R Walker, Krishnan Pillaipakkamnatt, Alexandra~M Davidson, Marysa Linares,
  and Domenick~J Pesce. 2019.
\newblock Automatic classification of rhetorical roles for sentences: Comparing
  rule-based scripts with machine learning.
\newblock In \emph{ASAIL@ ICAIL}.

\bibitem[{Wyner et~al.(2010)Wyner, Mochales-Palau, Moens, and
  Milward}]{wyner2010approaches}
Adam Wyner, Raquel Mochales-Palau, Marie-Francine Moens, and David Milward.
  2010.
\newblock Approaches to text mining arguments from legal cases.
\newblock In \emph{Semantic processing of legal texts}, pages 60--79. Springer.

\bibitem[{Wyner et~al.(2013)Wyner, Peters, and Katz}]{wyner2013case}
Adam~Z Wyner, Wim Peters, and Daniel Katz. 2013.
\newblock A case study on legal case annotation.
\newblock In \emph{JURIX}, pages 165--174.

\bibitem[{Xie et~al.(2020)Xie, Luong, Hovy, and Le}]{xie2020self}
Qizhe Xie, Minh-Thang Luong, Eduard Hovy, and Quoc~V Le. 2020.
\newblock Self-training with noisy student improves imagenet classification.
\newblock In \emph{Proceedings of the IEEE/CVF Conference on Computer Vision
  and Pattern Recognition}, pages 10687--10698.

\bibitem[{Yang et~al.(2019)Yang, Jia, Zhou, and Luo}]{ijcai2019-567}
Wenmian Yang, Weijia Jia, Xiaojie Zhou, and Yutao Luo. 2019.
\newblock \href {https://doi.org/10.24963/ijcai.2019/567} {Legal judgment
  prediction via multi-perspective bi-feedback network}.
\newblock In \emph{Proceedings of the Twenty-Eighth International Joint
  Conference on Artificial Intelligence, {IJCAI-19}}, pages 4085--4091.
  International Joint Conferences on Artificial Intelligence Organization.

\bibitem[{Ye et~al.(2018)Ye, Jiang, Luo, and Chao}]{ye-etal-2018-interpretable}
Hai Ye, Xin Jiang, Zhunchen Luo, and Wenhan Chao. 2018.
\newblock \href {https://doi.org/10.18653/v1/N18-1168} {Interpretable charge
  predictions for criminal cases: Learning to generate court views from fact
  descriptions}.
\newblock In \emph{Proceedings of the 2018 Conference of the North {A}merican
  Chapter of the Association for Computational Linguistics: Human Language
  Technologies, Volume 1 (Long Papers)}, pages 1854--1864, New Orleans,
  Louisiana. Association for Computational Linguistics.

\end{thebibliography}
\bibliographystyle{acl-natbib}

\clearpage
\newpage
\appendix
\section*{{\Large\selectfont{Appendix}}}

\section{Ethical Considerations} \label{app:ethics}

The proposed corpus and methods do not have direct ethical consequences to the best of our knowledge. The corpus is created from publicly available data from a public resource: \url{www.indiankanoon.org}. The website allows free downloads, and no copyrights were violated. With the help of law professors, we designed a course project centered around RR annotations for the student annotators. The students \textbf{voluntarily} participated in the annotations as a part of the course project. Moreover, annotators were curious about learning about AI technologies and further contributing towards its progress. There was no compulsion to take part in the annotation activity. 

The cases were selected randomly to avoid bias towards any entity, situation, or laws. Any meta-information related to individuals, organizations, and judges was removed so as to avoid any introduction of bias. For the application of corpus to judgment prediction task, we are not the first ones to do the task of judgment prediction. For the task, we took all the steps (names anonymization and removal of meta-information) as outlined in the already published work of \citet{malik-etal-2021-ildc}. The focus of this paper is rhetorical role prediction, and the task of judgment prediction is only a use-case. Moreover, in this paper we focus mainly on IT and CL cases where facts and scenarios are more objective and there are less biases compared to other types of cases (e.g., criminal and civil cases). As also described by \citet{malik-etal-2021-ildc}, we do not believe that the task could be fully automated, but rather it could augment the work of a judge or legal practitioner to expedite the legal process in highly populated countries. 

Legal-NLP is a relatively new area; we have taken all the steps to avoid any direct and foreseeable ethical implications; however, a lot more exploration is required by the research community to understand implicit ethical implications. For this to happen, resources need to be created, and we are making initial steps and efforts towards it. 

\section{Dataset and Annotations}

\subsection{Data Collection and Preprocessing} \label{app:preprocessing}

The IT and CL cases come from the Supreme Court of India, Bombay and Kolkata High Courts. For CL cases, we use the cases from the tribunals of NCLAT (National Company Law Appellate Tribunal)\footnote{\url{https://nclat.nic.in/}}, CCI (Competition Commission of India)\footnote{\url{https://www.cci.gov.in/}}, COMPAT (Competition Appellate Tribunal)\footnote{\url{http://compatarchives.nclat.nic.in}}. Since the IT laws are 50 years old and relatively dynamic, we stick to certain sections of IT domain only, whereas we use all the sections for CL domain. We restrict ourselves to the IT cases that are based on Section 147, Section 92C and Section 14A only to limit the subjectivity in cases. We randomly select 50 cases from IT and CL domain each to be annotated. We used regular expressions in Python to remove the auxillary information in the documents (For example: date, appellant and respondent names, judge names etc.) and filter out the main judgment of the document. We use the NLTK\footnote{\url{http://www.nltk.org/}} sentence tokenizer to split the document into sentences. The annotators were asked to annotate these sentences with the rhetorical roles. 

\subsection{Annotators Details}
With the help of law professors, we designed a course project centered around RR annotations for the student annotators. The students \textbf{voluntarily} participated in the annotations as a part of the course project. Moreover, annotators were curious about learning about AI technologies and further contributing towards its progress. There was no compulsion to take part in the annotation activity. 

The 6 annotators come from an Indian Law University. Three of them specialize in Income Tax domain and the other three specialize in Competition Law domain. 

\subsection{Rhetorical Roles} \label{app:roles}
\begin{figure}[h]
    \centering
    \includegraphics[width=0.5\textwidth]{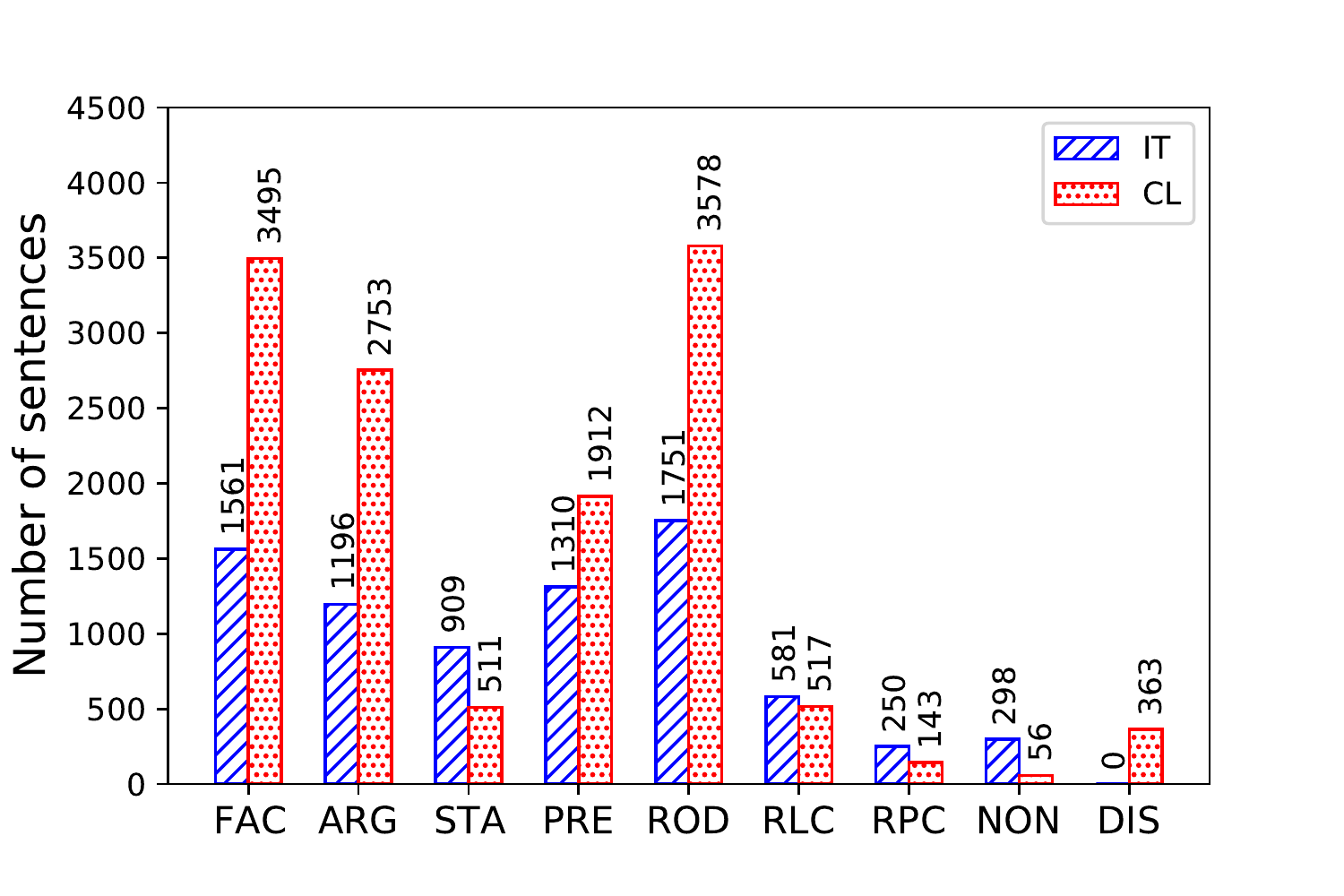}
    \caption{Distribution of RR labels in IT and CL documents.}
    \label{fig:CLstats}
\end{figure}

We provide the definition of each of the Rhetorical Role in the main paper. Examples for each of the RR are given in Table \ref{app:tab:rr-examples}. Figure \ref{fig:CLstats} provides the number of sentences for each label in the IT and CL dataset. Note that representation of both the domains is similar with the exception of DIS label.

\subsection{Secondary and Tertiary Annotation Labels} \label{app:secondary-tertiary}
Legal experts pointed out that a single sentence can sometimes represent multiple rhetorical roles (although this is not common). Each expert could also assign secondary and tertiary rhetorical roles to a single sentence to handle such scenarios and motivate future research. On an average annotators assigned secondary role in 5-7\% cases and assigned tertiary roles in 0.5-1\% cases. 

\subsection{Inter-annotator Agreement} \label{app:agreement}

\begin{figure}[]
     \centering
     \begin{subfigure}[b]{0.40\textwidth}
         \centering
         \includegraphics[width=\textwidth]{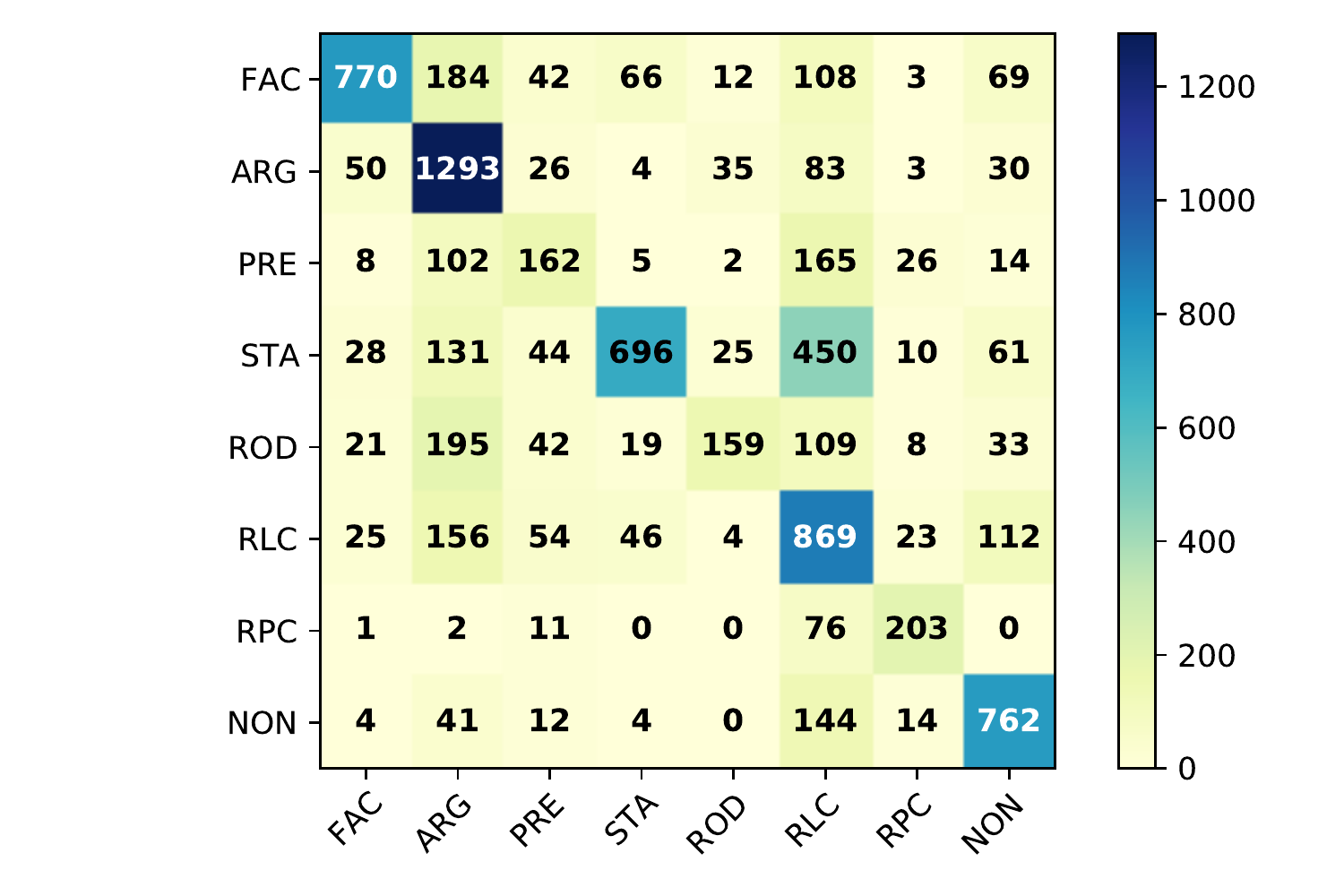}
         \caption{Between annotators $A_{1}$ and $A_{2}$ for IT domain}
         \label{fig:cm1}
     \end{subfigure}
     \hfill
     \begin{subfigure}[b]{0.40\textwidth}
         \centering
         \includegraphics[width=\textwidth]{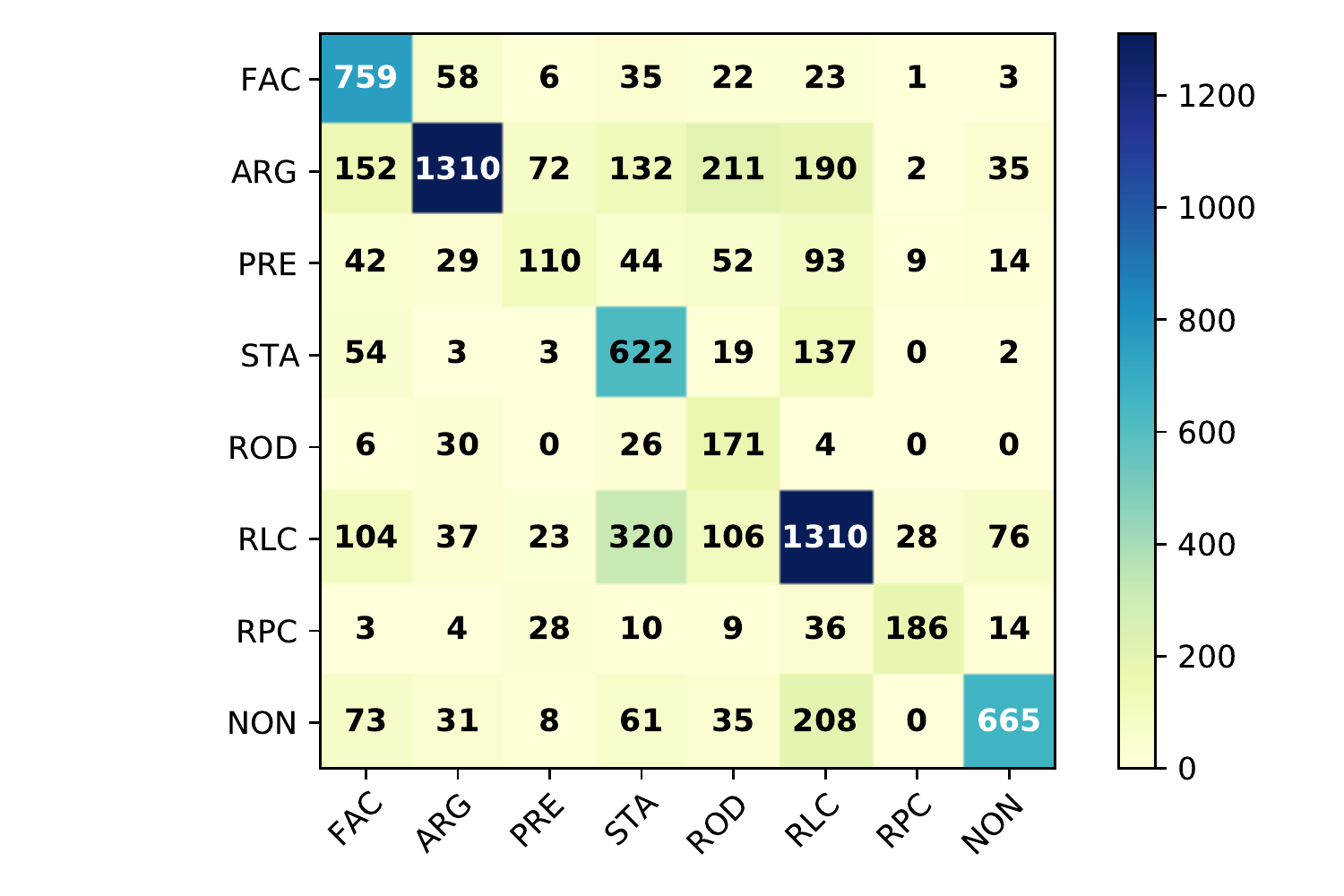}
         \caption{Between annotators $A_{2}$ and $A_{3}$ for IT domain}
         \label{fig:cm2}
     \end{subfigure}
    \begin{subfigure}[b]{0.40\textwidth}
         \centering
         \includegraphics[width=\textwidth]{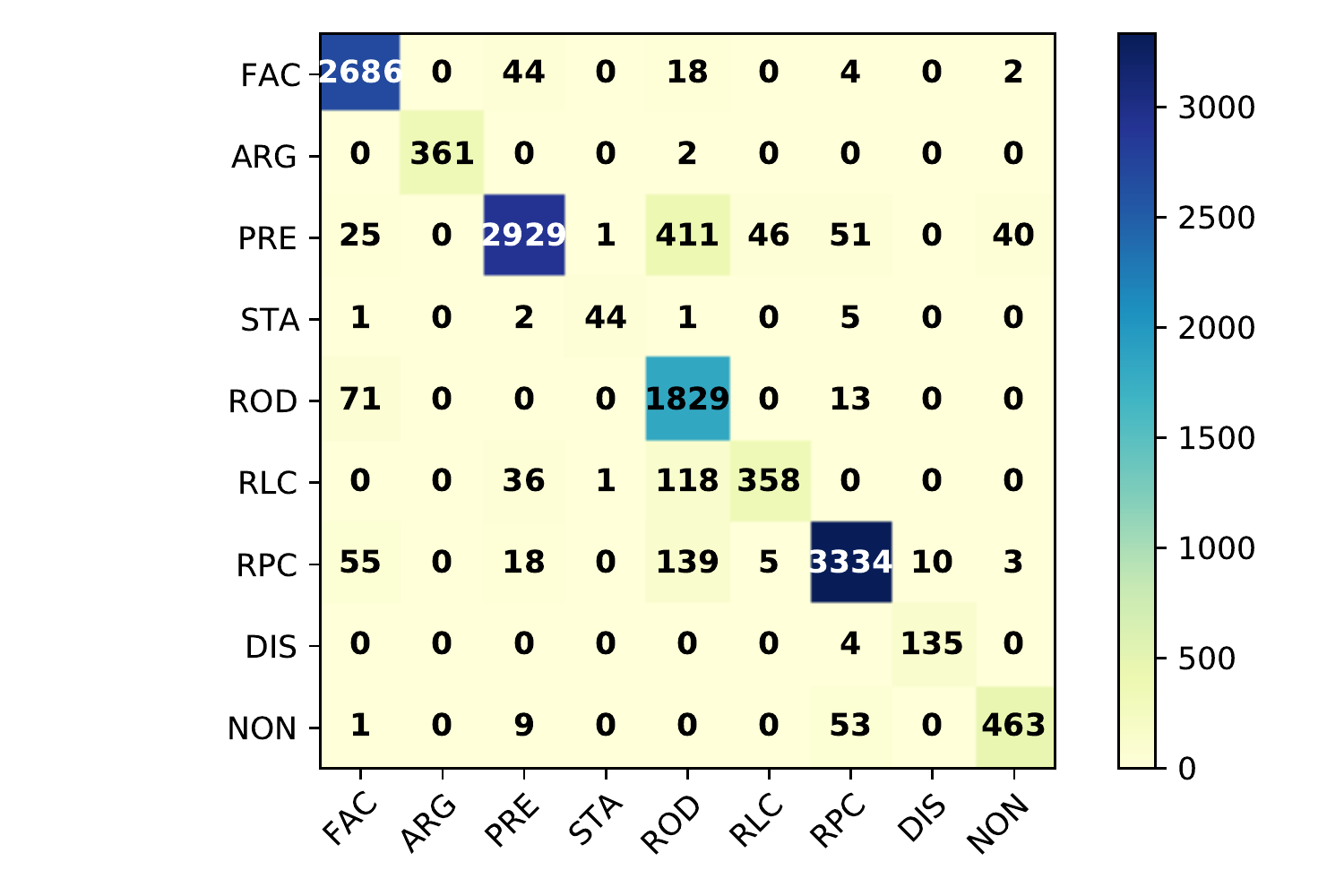}
         \caption{Between annotators $A_{1}$ and $A_{2}$ for CL domain}
         \label{fig:cm3}
     \end{subfigure}
    \begin{subfigure}[b]{0.40\textwidth}
         \centering
         \includegraphics[width=\textwidth]{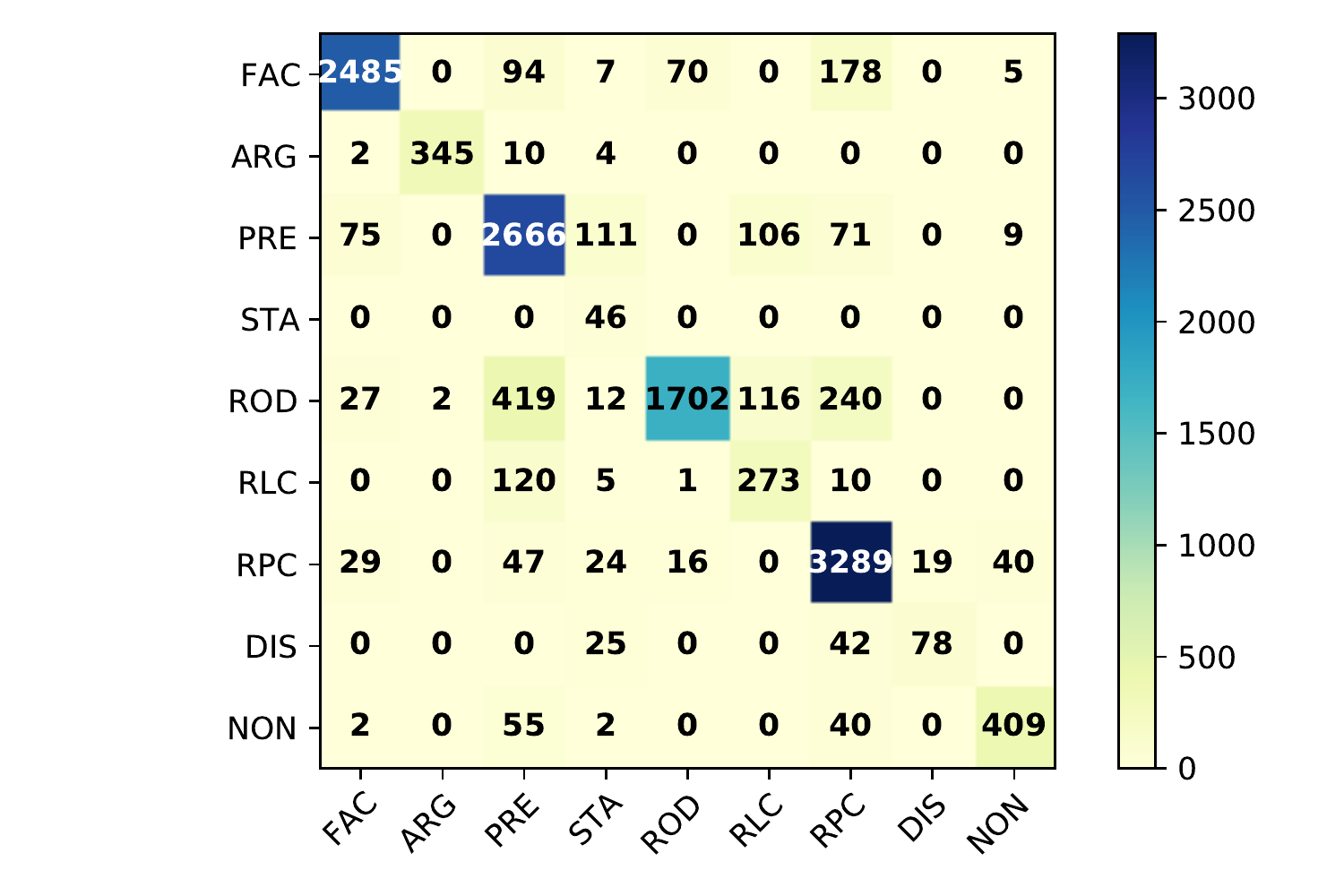}
         \caption{Between annotators $A_{2}$ and $A_{3}$ for IT domain}
         \label{fig:cm4}
     \end{subfigure}
     \hfill
        \caption{Confusion matrix between Annotators for IT and CL domains.}
        \label{fig:confusion-mat-remaining}
\end{figure}

Fleiss Kappa between all (fine-grained) labels is 0.59 for IT and 0.87 for CL, indicating substantial agreement. We provide the inter-annotator agreement (averaged pairwise macro F1 between annotators) upon 13 fine-grained labels in Table \ref{tab:interanno_pairwise_13labels}. Also, we provide the pairwise confusion matrices of annotators $(A_{1}, A_{2})$ and $(A_{2},A_{3})$ for both IT and CL domain in Figure \ref{fig:confusion-mat-remaining}.

\begin{table}[h]
\centering
\begin{tabular}{|c|c|c|}
\hline
\textbf{Label}       & \textbf{IT}   & \textbf{CL}   \\ \hline
\textbf{ARG-P}       & 0.74          & 0.90          \\ \hline
\textbf{ARG-R}       & 0.73          & 0.97          \\ \hline
\textbf{FAC}         & 0.77          & 0.88          \\ \hline
\textbf{ISS}         & 0.75          & 0.75          \\ \hline
\textbf{PRE-RU}      & 0.67          & 0.86          \\ \hline
\textbf{PRE-NR}      & 0.58          & 0.80          \\ \hline
\textbf{PRE-O}       & 0.43          & \_            \\ \hline
\textbf{STA}         & 0.78          & 0.89          \\ \hline
\textbf{RLC}         & 0.58          & 0.74          \\ \hline
\textbf{RPC}         & 0.75          & 0.74          \\ \hline
\textbf{ROD}         & 0.64          & 0.93          \\ \hline
\textbf{DIS}         & \_            & 0.98          \\ \hline
\textbf{NON}         & 0.45          & 0.52          \\ \hline
\textit{\textbf{F1}} & \textit{0.73} & \textit{0.88} \\ \hline
\end{tabular}
\caption{Label-wise inter-annotator agreement for all 13 fine-grained labels.}
\label{tab:interanno_pairwise_13labels}
\end{table}

\subsection{Annotation Analysis} \label{app:annotation-analysis} 

Annotation of judgments in order to identify and distinguish between the rhetorical roles played by its various parts is in itself a challenging task even for legal experts. We provide some qualitative examples of sentences and their corresponding rhetorical roles in Table \ref{app:tab:rr-examples} There are several factors involved in the exercise that requires the annotator to retrace the judicial decision making and recreate the impact left by the inputs available to the judge such as certain specific facts of the case, a particular piece of argument advanced by the lawyer representing one of the parties, or a judicial precedent from a higher court deemed applicable in the current case by the lawyer(s) or by the judge or by both. Moreover, the annotator only has access to the current document which is secondary account of what actually happened in the court. These limitations certainly makes the task of the annotator further difficult, and leaves them with no choice other than to make certain educated guesses when it comes to understanding the various nuances, both ostensible and probable, of certain rhetorical roles. It should, however, be noted that such variation need not occur for every rhetorical role, since not all the roles are equally susceptible to it –for instance, the facts of the case as laid down by the judge are more readily and objectively ascertainable by more than one annotator, whereas the boundaries between the issues framed by the judge and those deemed relevant as per the arguments advanced by the lawyers may blur more, especially because if the judge happens to agree with one of the lawyers and adopts their argument as part of the judicial reasoning itself. Similarly, it should also be noted that despite differing in their views of the nature and extent of rhetorical role played by a certain part of the judgment, the annotators may still agree with each other when it comes to identifying and segregating the final ruling made by the judge in that case –this phenomenon of having used two different routes to arrive at the same destination is not uncommon in the reenactment or ex-post-facto analysis of a judicial hearing and decision making process. A cumulative effect of the aforementioned factors can be observed in the results of the annotation. The analysis provided by the three annotators in case of competition law bear close resemblance with each other. On the other hand, in case of income tax law, the analysis provided by Users 1 and 3 bear greater resemblance with each other, compared to the resemblance between Users 1 and 2, or between Users 2 and 3. On a different note, it is also observed that the rhetorical role where the annotators have differed between themselves the most has been the point of Ruling made by the Lower Court, followed by the Ratio. This also ties in with the aforesaid argument that all rhetorical roles are not equally susceptible to the variation caused by the varying levels of success achieved by the different annotators in retracing the judicial thought pattern.

\subsection{Annotation Case Studies} \label{app:annotation-case-study} 

Along with law professors, we analyzed some of the case documents. Please refer to data files for the actual judgment. 

In the case of CL cases, the best resemblance that has been achieved is in the case of SC\_Competition Commission of India vs Fast Way Transmission Pvt Ltd and Ors 24012018 SC.txt, one would find that the judgment has been written in a manner as to provide specific indicators before every rhetorical role. For instance, before the Ruling by Lower Court starts, reference has been made that this is the opinion given the Competition Commission of India (the lower court in the relevant domain). Similarly, before Arguments made by Petitioner/Respondent, reference has been made that this is the argument made by the lawyer representing the petitioner/respondent. This judgment also provides a nice, consistent flow following the arrangement of the rhetorical roles in order. The relatively smaller size of the judgment also indicates a lower level of complexity (although there need not always be a consistent correlation between the two). 
On the other hand, if one considers the least resemblance achieved in the competition law domain, in the case of SC\_Excel Crop Care Limited vs Competition Commission of India and Ors 08052017 SC(1).txt, one would find that such specific indicators are usually absent, thus leaving scope for individual discretion and interpretation, the judgment goes back and forth between certain rhetorical roles (Issue, Ruling by Lower Court, Ratio by Present Court, Argument by Petitioner/Respondent, Precedent Relied Upon), and the relatively bigger size also involves additional complexity and analysis, which make room for further nuances as described above. 

Similarly, if one considers the best resemblance that has been achieved in the income tax domain, in the case of SC\_2014\_17.txt, one would find the case has involved fewer rhetorical roles, cut down on facts (mainly dealing with procedural issues on an appellate stage), and even among the rhetorical roles, it has focused on statutes and provisions thereof and the ratio and ruling. This has significantly reduced the possibility of the aforementioned richer jurisprudence, greater range of precedents, and resulting greater degree of subjective interpretation being at play. 
On the other hand, if one considers the least resemblance that has been achieved in the income tax domain, in the case of SC\_2008\_1597.txt, discusses Precedents to a greater detail including facts thereof, goes back and forth between certain rhetorical roles instead of maintaining a consistent order, and is not very clear about whether the judge is at times merely reiterating the arguments made by the lawyers, or is demonstrating their own view of such arguments. Collectively, these leave the scope for greater involvement of subjective interpretation of the aforesaid nuances. 

Yet on an overall basis, the elements of subjectivity, personal discretionary interpretation, and arbitrariness have been minimized by the selection of the chosen domains, along with the methodology adopted for annotation, thus leading to the present success attained in identification of rhetorical roles and using the same for prior relevant case identification and prediction.  

\section{Evaluation Metrics} \label{app:metrics}
We use the Macro F1 metric to evaluate the performance of models upon the task of Rhetorical Role labelling. Macro F1 is the mean of the label-wise F1 scores for each label. Given the true positives ($TP$), false positives ($FP$) and false negatives ($FN$), the F1 score for a single label is calculated as:
\begin{equation}
    F1 = \frac{TP}{TP + \left(\frac{FP+FN}{2}\right)}
\end{equation}

The pairwise inter-annotator agreement F1 between two annotators $A$ and $B$ is calculated by considering the annotations by annotator $A$ as the true labels and the annotations by annotator $B$ as the predicted labels. 

We also calculate Fleiss Kappa\footnote{https://en.wikipedia.org/wiki/Fleiss\%27\_kappa} to measure the inter-annotator agreement. 

\section{Model Training Details} \label{app:model}
All of our baseline experiments and training of Label shift prediction models (SBERT and BERT-SC) were conducted on Google Colab\footnote{https://colab.research.google.com/} and used the default single GPU Tesla P100-PCIE-16GB, provided by Colab. Our models were trained upon a single 11GB GeForce RTX 2080 TI. We used the SBERT model provided in the sentence-transformers library\footnote{https://pypi.org/project/sentence-transformers/}. We use the Huggingface\footnote{https://huggingface.co/} implementations of BERT-base and LEGAL-BERT models. Refer to Table \ref{tab:it_full}, \ref{tab:itcl_full} and \ref{tab:cl_full} for dataset-wise results and hyperparameters for each model. We also provide the training time and number of parameters of each model in Table \ref{tab:params_train_time}.

For SBERT-Shift, we kept the SBERT model as fixed and tuned the 3 linear layers on top. We used the Binary Crossentropy loss function with Adam Optimizer to tune the model upon the LSP task.

For BERT-SC, we fine-tuned the pre-trained BERT-base model upon the LSP task. We used the maximum sequence length of 256 tokens, a learning rate of $2e-5$ and kept the number of epochs as 5 during training. We used the same loss function and optimizer as the SBERT-Shift model.

\subsection{Reduced Label Set} \label{app:reducedLabels}

Due to the complexity of the task of RR prediction, we consider seven main labels (FAC, ARG, PRE, ROD, RPC, RLC, and STA) only. We plan to explore developing predictive models using fine-grained labels. 

\noindent \textbf{NON Label:} We ignore sentences with NON (None) labels (about 4\% for IT and 0.5\% for CL). We believe that this was necessary since the inter-annotator agreement for the NON label in both IT and CL domains, has an F1 score as low as 0.45, implying that even the legal experts themselves do not agree whether a particular sentence has a NON label. 

\noindent \textbf{Dissent Label:} Analysis of the annotated dataset reveals that the IT domain does not have any instance of dissent (DIS) label. There were only three documents (out of 50) in the CL domain having few instances of dissent label. Moreover, the instances of dissent label were present as a contiguous chunk of sentences at the end of the document. Hence, we discarded the sentences with dissent labels. Furthermore, law experts told us that the dissent phenomenon is rare; from a practical (application) point of view, these labels can be discarded.

\subsection{Single Sentence Classification Baselines}
We train single sentence classification models for the task of rhetorical role labelling. We use BERT-base-uncased and Legal-BERT models and fine-tune them upon the sentence classification task. We also try a variant of using context sentences (left sentence and the right sentence) along with the current sentence to make classification, we call this method BERT-neighbor. We use CrossEntropyLoss as the criterion and Adam as the optimizer. We use a batch size of 32 with a learning rate of 2e-5 and fine-tune for 5 epochs for all our experiments. Refer to Tables \ref{tab:it_full} , \ref{tab:cl_full} and \ref{tab:itcl_full} and for results and more information about the hyperparameters. 

\subsection{Sequence Classification Baselines}
We experiment with Sequence Classification Baselines like CRF with handcrafted features, BiLSTM with sent2vec embeddings and different versions of BiLSTM-CRF in which we varied the input embeddings. We experimented with sent2vec embeddings fine-tuned on Supreme Court Cases of India (same as in \cite{bhattacharya2019identification}). We also tried with sentence embeddings obtained from the BERT-base model. In another experiment, we fine-tuned a pre-trained BERT model upon the task of Masked Language Modelling (MLM) on the unlabelled documents of IT and CL domain, and used this model to extract the sentence embeddings for the BiLSTM-CRF model.

We used the same implementation of BiLSTM-CRF from \cite{bhattacharya2019identification}, with Adam optimizer and NLL loss function. Refer to Tables \ref{tab:it_full} , \ref{tab:cl_full} and \ref{tab:itcl_full} for experiment-wise hyperparameters.

\subsection{LSP-BiLSTM-CRF and MTL-BiLSTM-CRF models}
In our proposed approach of LSP-BiLSTM-CRF, we experiment with two methods of generating shift embeddings, namely BERT-SC and SBERT-Shift. These embeddings were then used as input to train a BiLSTM-CRF with similar training schedules. Refer to Tables \ref{tab:it_full} , \ref{tab:cl_full} and \ref{tab:itcl_full} for other hyperparameters.

\begin{table*}[]
\small
\centering
\begin{tabular}{|l|l|c|}
\hline
\textbf{Model}          & \textbf{\begin{tabular}[c]{@{}l@{}}Hyperparameters(E=Epochs),\\ (LR=Learning rate),\\ (BS=Batch Size),\\ (Dim=Embedding dimension),\\ (E1=Embedding dimension Shift),\\ (E2=Embedding dimension RR),\\ (H=Hidden dimension),\end{tabular}} & \multicolumn{1}{l|}{\textbf{IT (Macro F1)}} \\ \hline
BERT                    & LR=2e-5, BS=32, E=5                                                                                                                                                                                                                        & 0.56                                           \\ \hline
BERT-neighbor           & LR=2e-5, BS=32, E=5                                                                                                                                                                                                                        & 0.53                                           \\ \hline
Legal-BERT              & LR=2e-5, BS=32, E=5                                                                                                                                                                                                                        & 0.55                                           \\ \hline
CRF(handcrafted)        & LR=0.01, BS=40, Dim=172, E=300                                                                                                                                                                                                             & 0.55                                           \\ \hline
BiLSTM(sent2vec)        & LR=0.01, BS=40, Dim=200, H=100, E=300                                                                                                                                                                                                      & 0.55                                           \\ \hline
BiLSTM-CRF(handcrafted) & LR=0.01, BS=40, Dim=172, H=86, E=300                                                                                                                                                                                                       & 0.57                                           \\ \hline
BiLSTM-CRF(sent2vec)    & LR=0.01, BS=40, Dim=200, H=100, E=300                                                                                                                                                                                                      & 0.59                                           \\ \hline
BiLSTM-CRF(BERT emb)    & LR=0.01, BS=40, Dim=768, H=384, E=300                                                                                                                                                                                                      & 0.63                                           \\ \hline
BiLSTM-CRF(MLM emb)     & LR=0.01, BS=40, Dim=768, H=384, E=300                                                                                                                                                                                                      & 0.58                                           \\ \hline
LSP(SBERT)              & LR=0.005, BS=40, Dim=2304, H=1152, E=300                                                                                                                                                                                                   & 0.64                                           \\ \hline
LSP(BERT-SC)            & LR=0.005, BS=40, Dim=2304, H=1152, E=300                                                                                                                                                                                                   & 0.65                                           \\ \hline
MTL(MLM emb)            & \begin{tabular}[c]{@{}l@{}}LR=0.005, BS=40, E1=2304, E2=768 , H=1152(Shift), \\ H=384(RR),  E=300\end{tabular}                                                                                                                             & 0.67                                           \\ \hline
MTL(BERT-SC)            & \begin{tabular}[c]{@{}l@{}}LR=0.005, BS=40, E1=2304, E2=768, H=1152(Shift), \\ H=384(RR),  E=300\end{tabular}                                                                                                                              & 0.70                                           \\ \hline
MTL(BERT-SC)            & \begin{tabular}[c]{@{}l@{}}LR=0.005, BS=40, E1=2304, E2=2304, H=1152(Shift), \\ H=384(RR), E=300\end{tabular}                                                                                                                              & 0.68                                           \\ \hline
MTL(BERT-SC)            & \begin{tabular}[c]{@{}l@{}}LR=0.005, BS=40, E1=768, E2=768, H=1152(Shift), \\ H=384(RR),  E=300\end{tabular}                                                                                                                               & 0.64                                           \\ \hline
\end{tabular}
\caption{Hyperparameters and results on the IT dataset}
\label{tab:it_full}
\end{table*}

For MTL models, we experimented with different encoders $E_{1}$ and $E_{2}$. We experimented with using Shift embeddings (or BERT embeddings of sentences obtained from pre-trained BERT model) from BERT-SC in both the components. However, the best performing model was the one in which we used shift embeddings for the shift component and BERT embeddings for the RR component. We used the NLL loss in both components of the MTL model weighted by the hyperparameter $\lambda$. We use the Adam Optimizer for training. We provide dataset-wise hyperparameters and results in Tables \ref{tab:it_full} , \ref{tab:cl_full} and \ref{tab:itcl_full}. 

\begin{table*}[]
\small
\centering
\begin{tabular}{|l|l|c|}
\hline
\textbf{Model}          & \textbf{\begin{tabular}[c]{@{}l@{}}Hyperparameters(E=Epochs),\\ (LR=Learning rate),\\ (BS=Batch Size),\\ (Dim=Embedding dimension),\\ (E1=Embedding dimension Shift),\\ (E2=Embedding dimension RR),\\ (H=Hidden dimension),\end{tabular}} & \multicolumn{1}{l|}{\textbf{IT+CL (Macro F1)}} \\ \hline
BiLSTM-CRF(sent2vec)    & LR=0.01, BS=40, Dim=200, H=100, E=300                                                                                                                                                                                                      & 0.65                                           \\ \hline
BiLSTM-CRF(BERT)        & LR=0.01, BS=40, Dim=768, H=384, E=300                                                                                                                                                                                                      & 0.63                                           \\ \hline
LSP-BiLSTM-CRF(BERT-SC) & LR=0.005, BS=20, Dim=2304, H=1152, E=300                                                                                                                                                                                                   & 0.67                                           \\ \hline
MTL-BiLSTM-CRF(BERT-SC) & \begin{tabular}[c]{@{}l@{}}LR=0.005, BS=20, E1=2304, E2=768, \\ H=1152(Shift), H=384(RR),  E=300\end{tabular}                                                                                                                              & 0.70                                           \\ \hline
MTL-BiLSTM-CRF(BERT-SC) & \begin{tabular}[c]{@{}l@{}}LR=0.005, BS=20, E1=2304, E2=2304, \\ H=1152(Shift), H=384(RR),  E=300\end{tabular}                                                                                                                             & 0.68                                           \\ \hline
MTL-BiLSTM-CRF(BERT-SC) & \begin{tabular}[c]{@{}l@{}}LR=0.005, BS=20, E1=768, E2=768, \\ H=1152(Shift), H=384(RR),  E=300\end{tabular}                                                                                                                               & 0.65                                           \\ \hline
\end{tabular}
\caption{Hyperparameters and results on the combined (IT+CL) dataset}
\label{tab:itcl_full}
\end{table*}

\begin{table*}[]
\small
\centering
\begin{tabular}{|l|l|c|}
\hline
\textbf{Model}          & \textbf{\begin{tabular}[c]{@{}l@{}}Hyperparameters(E=Epochs),\\ (LR=Learning rate),\\ (BS=Batch Size),\\ (Dim=Embedding dimension),\\ (E1=Embedding dimension Shift),\\ (E2=Embedding dimension RR),\\ (H=Hidden dimension),\end{tabular}} & \multicolumn{1}{l|}{\textbf{CL (Macro F1)}} \\ \hline
BERT                    & LR=2e-5, BS=32, E=5                                                                                                                                                                                                                        & 0.52                                        \\ \hline
BERT-neighbor           & LR=2e-5, BS=32, E=5                                                                                                                                                                                                                        & 0.51                                        \\ \hline
Legal-BERT              & LR=2e-5, BS=32, E=5                                                                                                                                                                                                                        & 0.53                                        \\ \hline
CRF(handcrafted)        & LR=0.01, BS=40, Dim=172, E=300                                                                                                                                                                                                             & 0.52                                        \\ \hline
BiLSTM(sent2vec)        & LR=0.01, BS=40, Dim=200, H=100, E=300                                                                                                                                                                                                      & 0.54                                        \\ \hline
BiLSTM-CRF(handcrafted) & LR=0.01, BS=40, Dim=172, H=86, E=300                                                                                                                                                                                                       & 0.56                                        \\ \hline
BiLSTM-CRF(sent2vec)    & LR=0.01, BS=40, Dim=200, H=100, E=300                                                                                                                                                                                                      & 0.61                                        \\ \hline
BiLSTM-CRF(BERT emb)    & LR=0.01, BS=40, Dim=768, H=384, E=300                                                                                                                                                                                                      & 0.63                                        \\ \hline
BiLSTM-CRF(MLM emb)     & LR=0.01, BS=40, Dim=768, H=384, E=300                                                                                                                                                                                                      & 0.60                                        \\ \hline
LSP(SBERT)              & LR=0.005, BS=40, Dim=2304, H=1152, E=300                                                                                                                                                                                                   & 0.63                                        \\ \hline
LSP(BERT-SC)            & LR=0.005, BS=40, Dim=2304, H=1152, E=300                                                                                                                                                                                                   & 0.68                                        \\ \hline
MTL(MLM emb)            & \begin{tabular}[c]{@{}l@{}}LR=0.005, BS=20, E1=2304, E2=768 , H=1152(Shift), \\ H=384(RR),  E=300\end{tabular}                                                                                                                             & 0.67                                        \\ \hline
MTL(BERT-SC)            & \begin{tabular}[c]{@{}l@{}}LR=0.005, BS=20, E1=2304, E2=768, H=1152(Shift),\\  H=384(RR),  E=300\end{tabular}                                                                                                                              & 0.69                                        \\ \hline
MTL(BERT-SC)            & \begin{tabular}[c]{@{}l@{}}LR=0.005, BS=20, E1=2304, E2=2304, H=1152(Shift),\\  H=384(RR), E=300\end{tabular}                                                                                                                              & 0.67                                        \\ \hline
MTL(BERT-SC)            & \begin{tabular}[c]{@{}l@{}}LR=0.005, BS=20, E1=768, E2=768, H=1152(Shift),\\  H=384(RR),  E=300\end{tabular}                                                                                                                               & 0.64                                        \\ \hline
\end{tabular}
\caption{Hyperparameters and results on the CL dataset}
\label{tab:cl_full}
\end{table*}

\subsection{Hyperparameter $\lambda$} \label{app:sec-lambda}
We tuned the hyperparameter $\lambda$ of the MTL loss function upon the validation set. We trained the MTL model with $\lambda \in [0.1, 0.9]$ with strides of 0.1 and show the performance of our method on IT and IT+CL datasets in Figure \ref{app:fig:lambdavariation}. $\lambda= 0.6$ performs the best for the IT domain and also performs competitively on the combined domains.


\subsection{Model Distillation}
For model distillation experiments we trained the teacher model with same hyperparameters in Table \ref{tab:it_full} on the IT dataset. For the next two iteration of learning a student model, we used 48 unlabelled cases in each iteration. The weighing hyperparameter, $\alpha_{U}$ was kept as 0.3. In each iteration, the student model was trained with a batch size 16, a learning rate of 0.005 and for 300 epochs.

\begin{table*}[]
\centering
\begin{tabular}{|l|ll|ll|}
\hline
Model                & \multicolumn{2}{l|}{No of Parameters} & \multicolumn{2}{l|}{Training Time(min)} \\ \hline
                     & \multicolumn{1}{l|}{IT}          & CL         & \multicolumn{1}{l|}{IT}   & CL   \\ \hline
BiLSTM(sent2vec)     & \multicolumn{1}{l|}{240000}      & 240000     & \multicolumn{1}{l|}{15}        & 30        \\ \hline
BiLSTM-CRF(sent2vec) & \multicolumn{1}{l|}{240000}      & 240000     & \multicolumn{1}{l|}{15}        & 30        \\ \hline
BiLSTM-CRF(BERT emb) & \multicolumn{1}{l|}{3538944}     & 3538944    & \multicolumn{1}{l|}{30}        & 50        \\ \hline
BiLSTM-CRF(MLM emb)  & \multicolumn{1}{l|}{3538944}     & 3538944    & \multicolumn{1}{l|}{30}        & 50        \\ \hline
LSP(SBERT)           & \multicolumn{1}{l|}{31850496}    & 31850496   & \multicolumn{1}{l|}{90}        & 250       \\ \hline
LSP(BERT-SC)         & \multicolumn{1}{l|}{31850496}    & 31850496   & \multicolumn{1}{l|}{90}        & 250       \\ \hline
MTL(MLM emb)         & \multicolumn{1}{l|}{35411060}    & 35411060   & \multicolumn{1}{l|}{300}       & 1200      \\ \hline
MTL(BERT-SC)         & \multicolumn{1}{l|}{35411060}    & 35411060   & \multicolumn{1}{l|}{300}       & 1200      \\ \hline
\end{tabular}
\caption{Approx. number of parameters and computational budget of models.}
\label{tab:params_train_time}
\end{table*}

\begin{table*}[]
\centering
\begin{tabular}{|l|l|l|}
\hline
\textbf{Model} & \textbf{IT+CL docs}                             & \textbf{F1} \\ \hline
BERT-ILDC      & Predicted ROD \& RPC using BiLSTM-CRF(sent2vec) & 0.55        \\ \hline
BERT-ILDC      & Predicted ROD \& RPC using MTL(BERT-SC)         & 0.56        \\ \hline
\end{tabular}
\caption{Judgment Prediction results using predicted ROD \& RPC}
\label{app:tab:jp}
\end{table*}

\begin{table*}[]
\centering
\begin{tabular}{|l|l|}
\hline
Label                   & Sentence                                                                                                                                                                                                                                                                                                                        \\ \hline
Fact                    & \begin{tabular}[c]{@{}l@{}}It has also been alleged that the copies of the notices were also sent,\\ inter alia, to the principal officer of the said company and also to the ladies\\ as mentioned  herein before, who has sold the immovable property\\ in question.\end{tabular}                                              \\ \hline
Fact                    & \begin{tabular}[c]{@{}l@{}}For executing this contract, the assessee entered into various contracts \\-Offshore Supply contract and Offshore Service Contracts.\end{tabular}                                                                                                                                                   \\ \hline
Ruling By Lower Court   & \begin{tabular}[c]{@{}l@{}}But the words inland container depot were introduced in Section 2(12)\\ of the Customs Act, 1962, which defines customs port.\end{tabular}                                                                                                                                                          \\ \hline
Ruling By Lower Court   & \begin{tabular}[c]{@{}l@{}}We may also mention here that the cost of superstructure was \\Rs. 2,22,000 as per the letter of the assessee dated 28-11-66 addressed\\ to the ITO during the course of assessment proceedings.\end{tabular}                                                                                      \\ \hline
Argument                & \begin{tabular}[c]{@{}l@{}}Such opportunity can only be had by the disclosure of the materials to\\ the court as also to the aggrieved party when a challenge is thrown to the\\ very existence of the conditions precedent for initiation of the action.\end{tabular}                                                        \\ \hline
Argument                & \begin{tabular}[c]{@{}l@{}}In this connection, it was urged on behalf of the assessee(s) that, for the\\ relevant assessment years in question, the Assessing Officer was required\\ to obtain prior approval of the Joint Commissioner of Income Tax before\\ issuance of notice under Section 148 of the Act.\end{tabular} \\ \hline
Statute                 & \begin{tabular}[c]{@{}l@{}}In the meantime, applicant has to pay the additional amount of tax with\\ interest without which the application for settlement would not\\ be maintainable.\end{tabular}                                                                                                                             \\ \hline
Statute                 & \begin{tabular}[c]{@{}l@{}}On the other hand, interest for defaults in payment of advance tax falls\\ under section 234B, apart from sections 234A and 234C, in section\\ F of Chapter XVII.\end{tabular}                                                                                                                        \\ \hline
Ratio of  the Decision  & \begin{tabular}[c]{@{}l@{}}The State having received the money without right, and having retained\\ and used it, is bound to make the party good, just as an individual\\ would be under like circumstances.\end{tabular}                                                                                                     \\ \hline
Ratio of the Decision   & \begin{tabular}[c]{@{}l@{}}Therefore, the Department is right in its contention that under the\\ above situation there exists a Service PE in India (MSAS).\end{tabular}                                                                                                                                                       \\ \hline
Ruling by Present Court & \begin{tabular}[c]{@{}l@{}}For these reasons, we hold that the Tribunal was wrong in reducing the\\ penalty imposed on the assessee below the minimum prescribed\\ under Section 271(1)(iii) of the Income-tax Act, 1961.\end{tabular}                                                                                        \\ \hline
Ruling by Present Court & \begin{tabular}[c]{@{}l@{}}Hence, in the cases arising before 1.4.2002, losses pertaining to exempted\\ income cannot be disallowed.\end{tabular}                                                                                                                                                                              \\ \hline
Precedent               & \begin{tabular}[c]{@{}l@{}}Yet he none the less remains the owner of the thing, while all the\\ others own nothing more than rights over it.\end{tabular}                                                                                                                                                                      \\ \hline
Precedent               & \begin{tabular}[c]{@{}l@{}}I understand the Division Bench decision in Commissioner of\\ Income-tax v. Anwar Ali, only in that context.\end{tabular}                                                                                                                                                                           \\ \hline
None                    & Leave granted.                                                                                                                                                                                                                                                                                                                  \\ \hline
None                    & There is one more way of answering this point.                                                                                                                                                                                                                                                                                  \\ \hline
Dissent                 & Therefore a constructive solution has to be found out.                                                                                                                                                                                                                                                                          \\ \hline
Dissent                 & \begin{tabular}[c]{@{}l@{}}In the light of the Supreme Court decision in the case of CCI vs SAIL\\ (supra) t his issue has to be examined.\end{tabular}                                                                                                                                                                         \\ \hline
\end{tabular}
\caption{Example sentences for each label.}
\label{app:tab:rr-examples}
\end{table*}


    

\end{document}